\documentclass[journal]{IEEEtran}
\usepackage{subcaption}
\usepackage{amssymb}
\usepackage{multirow}
\usepackage{multicol}
\usepackage{color}
\usepackage{cite}
\usepackage{hyperref}
\usepackage{amsmath,amsfonts}
\usepackage{algorithm,algpseudocode}
\usepackage{array}
\usepackage{booktabs}
\usepackage{cleveref}
\usepackage{pifont}
\usepackage{makecell}
\usepackage{svg}
\usepackage{adjustbox}

\usepackage{textcomp}
\usepackage{stfloats}
\usepackage{url}
\usepackage{verbatim}
\usepackage{graphicx}
\def\BibTeX{{\rm B\kern-.05em{\sc i\kern-.025em b}\kern-.08em
    T\kern-.1667em\lower.7ex\hbox{E}\kern-.125emX}}
\usepackage{balance}

\begin{document}
\title{Open-world Point Cloud Semantic Segmentation:\\ A Human-in-the-loop Framework}

\author{Peng~Zhang,
	    Songru~Yang,
	    Jinsheng~Sun,
	    Weiqing~Li,
	    and~Zhiyong~Su 
\IEEEcompsocitemizethanks{\IEEEcompsocthanksitem P. Zhang, S. Yang, J. Sun and Z. Su are with School of Automation, Nanjing University of Science and Technology, Nanjing 210094, China.\protect\\
E-mails: \{zhangpeng, yangsongru\}@njust.edu.cn, jssun67@163.com, su@njust.edu.cn
\IEEEcompsocthanksitem W. Li is with School of Computer Science and Engineering, Nanjing University of Science and Technology, Nanjing 210094, China.\protect\\
E-mail: li\_weiqing@njust.edu.cn}
\thanks{Manuscript received April 19, 2005; revised August 26, 2015. (Corresponding author: Zhiyong Su.)}}

\markboth{Journal of \LaTeX\ Class Files,~Vol.~18, No.~9, September~2020}%
{Open-world Point Cloud Semantic Segmentation: A Human-in-the-loop Framework}

\maketitle

\begin{abstract}
Open-world point cloud semantic segmentation (OW-Seg) aims to predict point labels for both base and novel classes in real-world scenarios.
However, existing methods depend on resource-intensive offline incremental learning or pre-collected annotated support data, limiting their practicality.
Furthermore, the distribution shifts across support (training) samples and query (testing) samples introduce biased class prototypes, resulting in sub-optimal predictions.
To address these limitations, we propose HOW-Seg, the first human-in-the-loop framework for OW-Seg.
Instead of relying on additional annotated samples, HOW-Seg constructs class prototypes in the query sample feature space by sparse human-in-the-loop annotations, thereby avoiding the prototype bias caused by intra-class distribution shifts.
Considering the lack of granularity of initial prototypes, we introduce an interactive prototype disambiguation mechanism to refine ambiguous prototypes, which correspond to annotations of different classes.
To further enrich contextual awareness, we propose a prototype label assignment module, which employs a dense conditional random field (CRF) upon the prototypes to optimize their label assignments.
Through iterative human feedback, HOW-Seg dynamically improves its predictions, achieving high-quality segmentation for both base and novel classes.
Experiments demonstrate that with sparse annotations (e.g., one-class-one-click), HOW-Seg surpasses the state-of-the-art generalized few-shot segmentation (GFS-Seg) method under the 5-shot setting.
When using advanced backbones (e.g., Stratified Transformer) and denser annotations (e.g., 10 clicks), HOW-Seg achieves 85.27\% mIoU on S3DIS and 66.37\% mIoU on ScanNetv2, significantly outperforming other alternatives.
The source code will be publicly available at \href{https://github.com/Pengz98/HOW-Seg}{https://github.com/Pengz98/HOW-Seg}.
\end{abstract}


\section{Introduction}
\IEEEPARstart{P}{oint} cloud semantic segmentation has achieved significant progress over the past decade, driven by advancements in deep learning techniques \cite{choy-2019-cvpr,qi-2017-cvpr,wang-2019-tog,zhao-2021-iccv,lai-2022-cvpr,kolodiazhnyi-2024-cvpr}.
However, a considerable gap persists between research and practical applications. 
Most existing methods operate under the static closed-set assumption, which presumes that all possible classes are considered during training. 
As a result, these closed-set segmentors struggle in dynamic open-world scenarios where unknown novel classes frequently appear, misclassifying them as one of the known base classes.
This challenge highlights the need for open-world semantic segmentation (OW-Seg), where segmentors must handle both known base classes and unknown novel classes that may emerge during testing.

\begin{figure}[!t]
\centering
\includegraphics[width=0.5\textwidth]{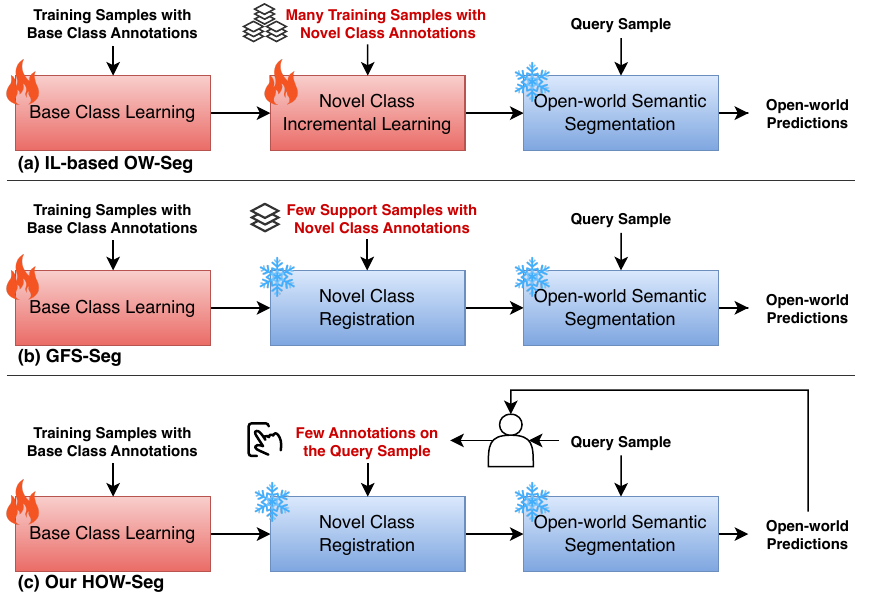}
\caption{Pipeline comparison of open-world point cloud semantic segmentation methods.
The common objective is to obtain open-world predictions encompassing both base and novel classes.
Three distinct pipelines are demonstrated:
\textbf{(a) Incremental Learning (IL)-based OW-Seg} requires computationally expensive fine-tuning with extensive novel class annotated samples;
\textbf{(b) Generalized Few-shot Semantic Segmentation (GFS-Seg)} leverages densely annotated support data for knowledge registration of the novel classes;
\textbf{(c) Our Human-in-the-loop OW-Seg (HOW-Seg)} directly registers novel class information with sparse query annotations, without additional fine-tuning stages or support samples.
We use red rectangles with flame icons to represent the offline training processes, and blue rectangles with snowflake icons to represent the online inference processes.}
\label{fig:intro}
\end{figure}

Current OW-Seg approaches are broadly categorized into two methodological pipelines: incremental learning (IL)-based OW-Seg methods \cite{cen-2022-eccv,xu-2024-cvpr} and generalized few-shot semantic segmentation (GFS-Seg) methods \cite{tian-2022-cvpr,xu-2023-iccv,tsai-2024-eccv}.
IL-based OW-Seg methods learn to segment novel classes by offline incremental learning with many annotated training samples, as shown in \Cref{fig:intro}(a).
The IL-based OW-Seg methods are executed sequentially across two learning stages: the base class learning stage and the novel class incremental learning stage.
During the base class learning stage, where only labels from a base class closed-set are available, the objective is to train an open-set model capable of predicting known base classes while identifying unknown classes out of the base class set.
In the novel class IL stage, where novel class labels are introduced, the goal is to incrementally update the model by incorporating knowledge from the novel class labels while retaining the ability to segment base classes.
The resulting open-world model is able to predict labels for both base and novel classes.
GFS-Seg is the other alternative for OW-Seg, which depends on few annotated support samples containing the novel classes occurring in the query sample, as shown in \Cref{fig:intro}(b).
Unlike IL-based OW-Seg updating novel class knowledge with abundant annotated training samples through offline incremental learning, GFS-Seg registers novel class information based on few annotated support samples during inference.

\begin{figure*}[ht]
\centering
\includegraphics[width=0.9\textwidth]{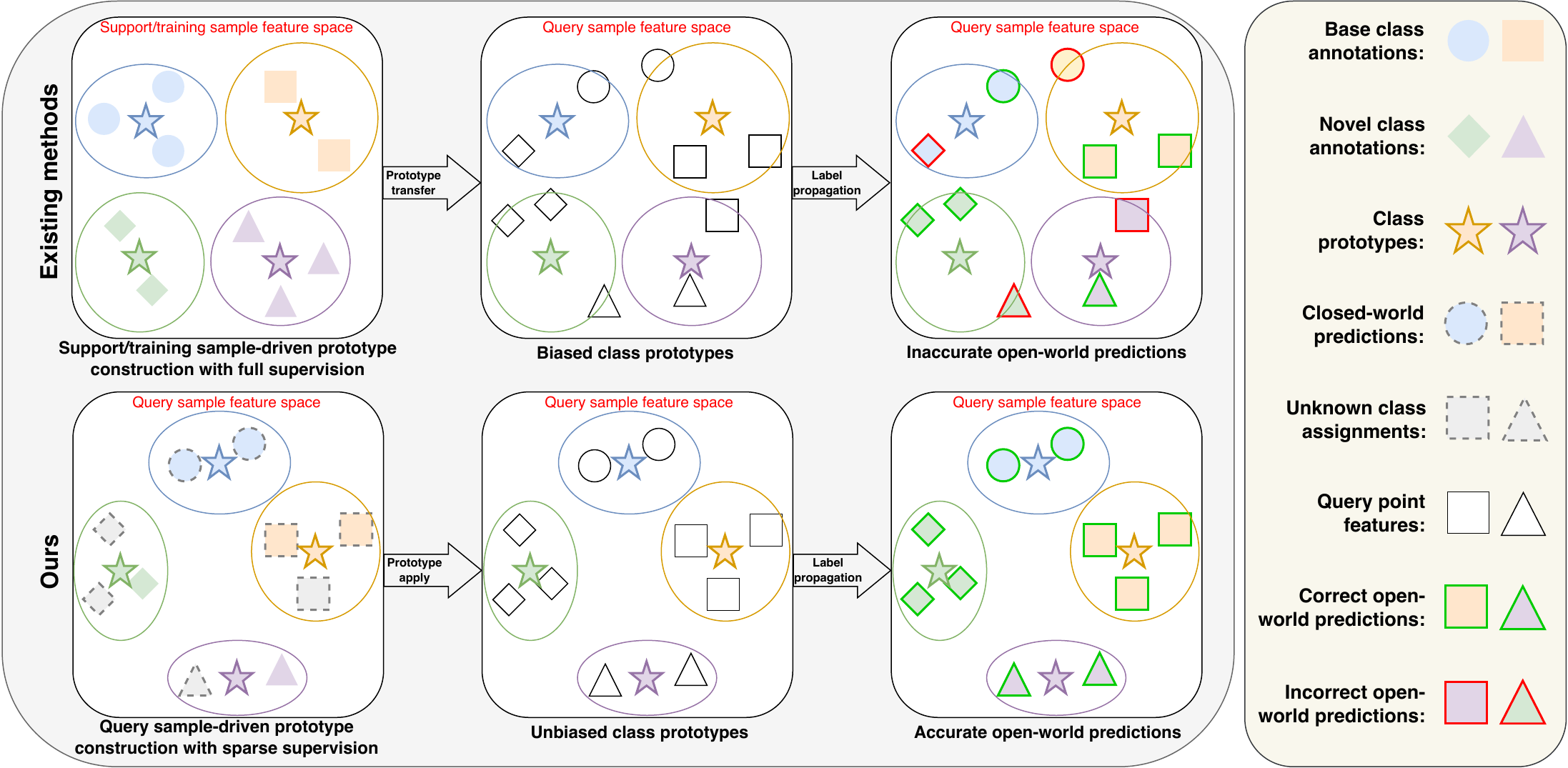}
\caption{
Existing methods (i.e., IL-based OW-Seg and GFS-Seg methods) transfer class prototypes across different samples (from support/training samples to the query sample), leading to biased class prototypes and inaccurate open-world predictions.
Our method applies class prototypes directly extracted from the current query sample based on sparse annotations and open-set predictions, enabling unbiased class prototypes and accurate open-world predictions.}
\label{fig:intro_b}
\end{figure*}

However, both pipelines rely on additional annotated samples, incurring limitations in practice.
IL-based OW-Seg methods demand computationally expensive offline fine-tuning with resource-intensive point-level annotations.
While GFS-Seg methods avoid offline incremental learning, they depend on pre-collected annotated support samples, increasing computation and memory overhead during inference.
Theses process must be repeated whenever novel classes emerge, hindering flexibility in dynamic open-world scenarios where novel classes frequently appear.
Furthermore, neither approach adequately addresses intra-class distribution shifts between support (training) samples and query samples, leading to biased novel class prototypes and degraded segmentation performance.
As shown in \Cref{fig:intro_b}, existing methods construct novel class prototypes based on annotated support or training samples.
When applied to the query sample, these class prototypes become biased in the query feature space due to cross-sample distribution shifts, resulting in inaccurate open-world predictions.

To address these limitations, we propose a novel \textbf{H}uman-in-the-loop \textbf{O}pen-\textbf{W}orld semantic \textbf{Seg}mentation (HOW-Seg) framework 3D point clouds. 
As illustrated in \Cref{fig:intro}, our approach eliminates the reliance on additional annotated samples, instead enabling prediction of labels for both base and novel classes using only sparse annotations on the query sample.
By leveraging human-in-the-loop guidance, HOW-Seg dynamically construct novel class prototypes directly in the query sample feature space, avoiding intra-class distribution shifts across samples (see \Cref{fig:intro_b}).
Specifically, we first obtain the point features and closed-world predictions with the segmentation backbone pre-trained in the closed-world fashion.
Then, we construct initial prototypes by K-Means \cite{lloyd-1982-tit} clustering on the point features augmented by closed-world predictions.
After that, we update the initial prototypes through the prototype disambiguation module to alleviate semantic conflicts within prototypes.
Next, we assign labels from sparse annotations to those prototypes by similarity matching.
Those labels are further propagated across prototypes based on a dense conditional random field (CRF), resulting in context-aware prototype label assignment.
Finally, we obtain the open-world point predictions by propagating the labels from prototypes to their associated points.
Notably, with progressive human annotations, the prototypes can be iteratively updated and labeled, resulting in evolving high quality segmentation results.
We evaluate our method with different segmentation backbones on the S3DIS \cite{armeni-2016-cvpr} and ScanNetv2 \cite{dai-2017-cvpr} datasets, showing superior performance upon the state-of-the-art GFS-Seg method with only one click for each class.
Our overall contributions are summarized as follows:
\begin{itemize}
\item We propose HOW-Seg, a novel human-in-the-loop framework for open-world point cloud semantic segmentation, eliminating the need of additional annotated data or learning stages.
\item The proposed HOW-Seg dynamically constructs class prototypes based on query point features alongside sparse annotations, avoiding prototype bias caused by intra-class distribution shifts across different samples.
\item We design an interactive prototype disambiguation module for discriminative prototype construction and a CRF-based prototype label assignment module for context-aware open-world predictions.
\item We conduct comprehensive experiments to show the effectiveness of HOW-Seg in high quality open-world semantic segmentation of point clouds.
\end{itemize}

\section{Related Works} \label{sec:related_works}
\subsection{Closed-world semantic segmentation for point clouds}
Closed-world point cloud semantic segmentation methods assume that all possible classes are considered during training and no more classes would emerge during inference.
Under this closed-world paradigm, segmentation models typically structured with a feature extractor and linear classifier, the latter configured with an output noses matching the number of pre-defined classes.
Based on the type of feature extracting, existing closed-world point cloud semantic segmentation models can be divided into projection-based, voxel-based and point-based models.
Projection-based \cite{lawin-2017-caip,boulch-2017-3dor,boulch-2018-cg} models project the point cloud into multi-view projection maps and extract pixel-wise features by 2D convolutional neural networks (CNN), which are aggregated to corresponding 3D points for point-wise label classification.
Voxel-based models \cite{wang-2017-tog,choy-2019-cvpr,zhang-2020-eccv,ye-2022-eccv,schult-2023-icra,peng-2024-cvpr} convert unstructured point clouds into structured voxels and employ 3D CNN to extract voxel-wise features, which are propagated to all points within the voxel for label classification.
Point-based models directly process the raw points, thereby improving efficiency and avoiding information loss.
Pioneering point-based models \cite{qi-2017-cvpr,qi-2017-nips,li-2018-nips,wang-2019-tog,yin-2023-tcsvt,zhao-2023-tcsvt} utilize multi-layer perceptrons (MLPs) to extract point-wise features and obtain the global features with symmetric aggregation.
Wang et al. \cite{wang-2019-tog} build local and non-local $k$-nearest neighbor ($k$-NN) graphs on 3D points and capture local contextual features through graph convolutional neural networks (GCNN).
Recent works \cite{zhao-2021-iccv,guo-2021-cvm,park-2022-cvpr,lai-2022-cvpr,wu-2022-nips,wang-2023-tog,wu-2024-cvpr,kolodiazhnyi-2024-cvpr} employ the transformer architecture in the point-based models considering its capability in capturing long-range contextual features.
We adopt the DGCNN \cite{wang-2019-tog} and stratified transformer (ST) \cite{lai-2022-cvpr} as the segmentation backbones considering their effectiveness and popularity in point cloud semantic segmentation.

Although the point cloud semantic segmentation models has witnessed remarkable development, the closed-world assumption hinders their practicality in real-world scenarios.
To address this practical issue, we propose to leverage sparse human annotations to enable closed-world segmentation models to predict labels for both base and novel classes.

\subsection{Incremental learning-based open-world semantic segmentation for point clouds}
Incremental learning (IL)-based point cloud open-world semantic segmentation (OW-Seg) methods solve the OW-Seg problem in two subsequent tasks, i.e., open-set semantic segmentation (OS-Seg) which identifies the unknown classes out of the known ones and incremental learning which incrementally update the models with novel class supervisions.
Cen et al. \cite{cen-2022-eccv} first propose a novel framework named REAL for LIDAR point clouds.
In the OS-seg task, they employ redundancy classifiers upon the original closed-set segmentation network to predict probabilities for base classes and the unknown class.
In the IL task, some redundancy classifiers are trained to recognize novel classes while others are still responsible for identifying the unknown classes.
Recently, Xu et al. \cite{xu-2024-cvpr} propose a probability-driven framework (PDF) for OW-Seg on indoor scene point clouds.
The PDF consists of a pseudo-labeling scheme for the OS-Seg task to capture features of unknown classes and an incremental knowledge distillation strategy for the CIL task to integrate novel classes into existing knowledge base.

However, those IL-based OW-Seg methods rely on the costly IL task, requiring computationally expensive fine-tuning processes with labor-intensive novel class annotations.
Furthermore, the incrementally learned segmentor may struggle with dynamic open-world scenarios due to the intra-class distribution shifts across samples.
Therefore, we propose involving humans in the segmentation loop, allowing them to provide sparse annotations on the current query data to construct unbiased novel class representations without relying on IL stages.

\subsection{Few-shot semantic segmentation for point clouds}
Point cloud few-shot semantic segmentation (FS-Seg) enables the prediction of novel classes by leveraging the support set containing examples of these classes with annotations.
AttMPTI \cite{zhao-2021-cvpr} is the pioneering work in point cloud FS-Seg, which extracts multiple prototypes on the support features and propagates the labels to query points by constructing $k$-NN graph in the feature space.
The multi-prototypes represent centroids in the feature space with certain semantics, typically acting as the minimum segmentation units for few-shot segmentation \cite{zhao-2021-cvpr,cong-2024-tmm,wei-2024-nips} and weakly supervised segmentation \cite{chen-2022-cvpr,su-2023-tcsvt,tang-2024-cvpr}.
A critical challenge is the intra-class distribution shifts between query and support samples, resulting in biased prototypes for query samples.
Subsequently works \cite{ning-2023-mm,he-2023-tip,lang-2023-tpami,zheng-2024-tcsvt,zhu-2024-tcsvt} improve the overall performance through feature optimization so as to enhance the correlations between query and support features.
Hu et al. \cite{hu-2024-tcsvt} propose a query-guided network to explore support prototypes, which leverages geometry relationship between query point features and prototypes to address data misalignment due to intra-class variations.
An et al. \cite{an-2024-cvpr} propose a correlation optimization method named COSeg that explicitly refines the correlations between support prototypes and query features through learnable modules.
Zhu et al. \cite{zhu-2024-cvpr} leverage non-parametric networks to narrow the feature gap between support and query samples.
However, these FS-Seg methods only focus on novel classes indicated by support samples, dissatisfying the requirement of OW-Seg.

Generalized few-shot semantic segmentation (GFS-Seg) extends FS-Seg to the OW-Seg scenario by registering novel class knowledge from support samples while maintaining base class knowledge during base class learning.
Tian et al. \cite{tian-2022-cvpr} first propose the GFS-Seg setting for images, and solve it by context-aware prototype learning (CAPL) that leverages co-occurrence prior knowledge from support data and enriches contextual information based on each query data.
Xu et al. \cite{xu-2023-iccv} first apply GFS-Seg to point clouds by incorporating geometric prototypes into the CAPL framework to facilitate generalization to novel classes while maintaining performance on the base classes.
Tsai et al. \cite{tsai-2024-eccv} assigns pseudo classes to unlabeled background to force the model to learn distinctive point features for separation of potential novel classes.
The pseudo labeling process is based on image and text encoders, requiring additional projection techniques and text prompts.

However, GFS-Seg highly relies on annotated support samples with novel classes occurring in the query sample, which might be too strong in practice considering the cumbersome data collection and annotation processes.
Furthermore, despite extensive efforts to strengthen the correlation between support and query features, cross-sample distribution shifts remain inevitable, resulting in biased class representations that hinder accurate open-world predictions.

\section{Method} \label{sec:method}
We propose a human-in-the-loop framework for open-world point cloud semantic segmentation.
In this section, we first formally define the human-in-the-loop open-world semantic segmentation problem for point clouds.
Next, we provide an overview of the proposed HOW-Seg framework, which is built on the prototype theory.
Then, we describe how prototypes are initialized, updated and labeled, acting as the minimum unit for open-world semantic segmentation.
Finally, we present how the open-world predictions are obtained with the constructed prototypes and human sparse annotations.

\begin{figure*}[t]
\centering
\includegraphics[width=1.0\textwidth]{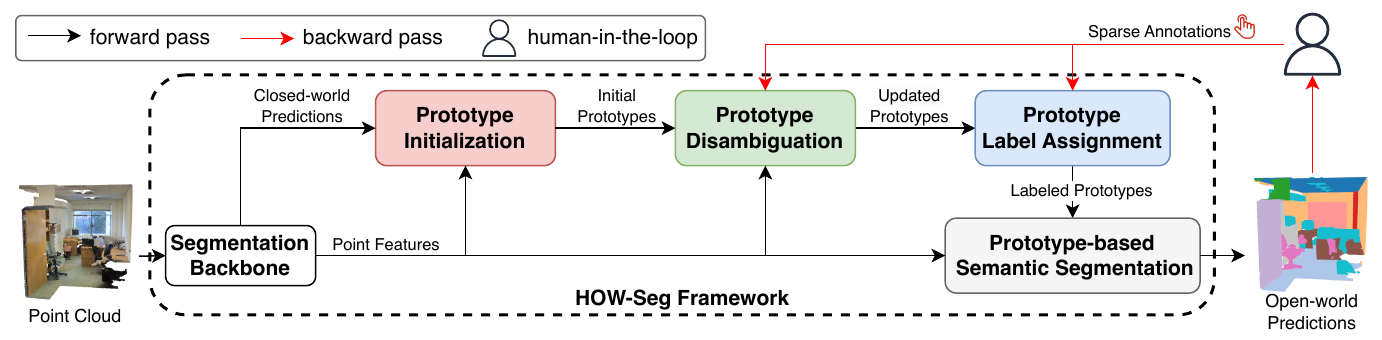}
\caption{
Overview of the proposed HOW-Seg framework.
Given a segmentation backbone pre-trained in the closed-world fashion, our framework enables open-world point cloud semantic segmentation with human sparse annotations following the steps:
First, we construct the initial prototypes based on point features and closed-world predictions from the segmentation backbone;
Second, we identify and reconstruct ambiguous prototypes based on human sparse annotations, resulting in updated prototypes;
Third, we assign labels for all  prototypes based on limited human annotations;
Finally, we conduct prototype-based semantic segmentation based on those labeled prototypes, obtaining the open-world predictions on point features.}
\label{fig:method_overview}
\end{figure*}

\subsection{Problem definition}
We aim to solve open-world point cloud semantic segmentation in a human-in-the-loop way.
A training dataset $D_{train}$ with label space $C_{train}=C_b$ and a testing dataset $D_{test}$ with label space $C_{test}=C_b \cup C_n$ are provided, where $C_b$ and $C_n$ denote the base and novel classes ($C_b\cap C_n = \varnothing$), respectively.
For each point cloud in the $D_{train}$, we label points that do not belong to any of $C_b$ as the unknown class, following the setting of  \cite{chakravarthy-2024-ijcv}.
We train a segmentor $S_{\theta}(.): \mathbb{R}^d \rightarrow \mathbb{R}^{(|C_b|+1)}$ on the $D_{train}=\{(\mathbf{P}_i,\mathbf{M}_i)\}_{i=1}^{|D_{train}|}$ in a common supervised manner, where each point cloud $\mathbf{P}_i \in \mathbb{R}^{n\times d_0}$ consists of $n$ points with feature dimension $d_0$, $\mathbf{M}_i$ denotes the point-wise annotation of $(|C_b| + 1)$ classes, and the $(|C_b| + 1)^{th}$ class denotes the $unknown$ class.
The segmentor $S_{\theta}(.)$ can be further decomposed into a feature extractor $\text{FE}(.): \mathbb{R}^d \rightarrow \mathbb{R}^{f}$ and a linear classifier $\text{Lin}(.): \mathbb{R}^f \rightarrow \mathbb{R}^{|C_b| + 1}$, where $f$ denotes the dimension of the extracted features.

With the base class training on the $D_{train}$ mentioned above, the segmentor learns good representations that can separate different base classes and the unknown class, which is expected to be generalized to different novel classes.
Given the pre-trained segmentor, our goal is to apply it to the $D_{test}=\{P_i\}_{i=1}^{|D_{test}|}$, producing dense predictions of both base and novel classes with sparse annotations provided.

\subsection{HOW-Seg framework}
\Cref{fig:method_overview} illustrates our HOW-Seg framework, mainly consisting of four components: the prototype initialization module, the prototype disambiguation module and the prototype label assignment module and the prototype-based semantic segmentation module.
Given a point cloud containing novel classes, we first extract point features using the feature extractor of segmentation backbone.
Then, the prototype initialization module constructs initial prototypes.
After that, the prototype disambiguation module checks the ambiguity of each constructed prototype and updates the ambiguous ones based on human sparse annotations.
Next, the prototype label assignment module assigns labels for the updated prototypes based on human sparse annotations and prototype-level contexts.
Finally, the prototype-based semantic segmentation module propagates the labels from prototypes to their corresponding points based on cosine similarity, resulting in open-world predictions.
If necessary, one can further provide additional annotations to drive prototype update and label re-assignment, progressively refining the predictions.

\begin{figure}[!t]
\centering
\includegraphics[width=0.48\textwidth]{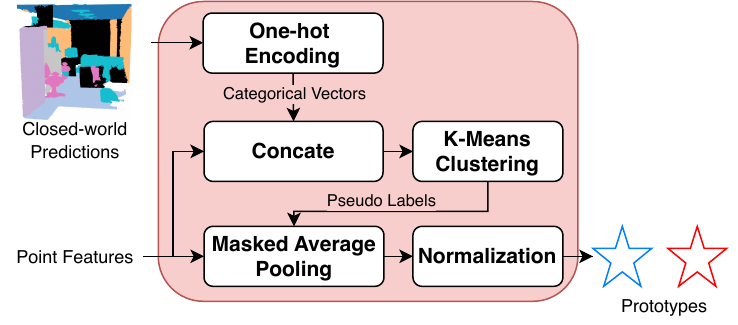}
\caption{
Workflow of prototype initialization.}
\label{fig:method_prototype_initialization}
\end{figure} 

\begin{figure*}[!t]
\centering
\includegraphics[width=0.9\textwidth]{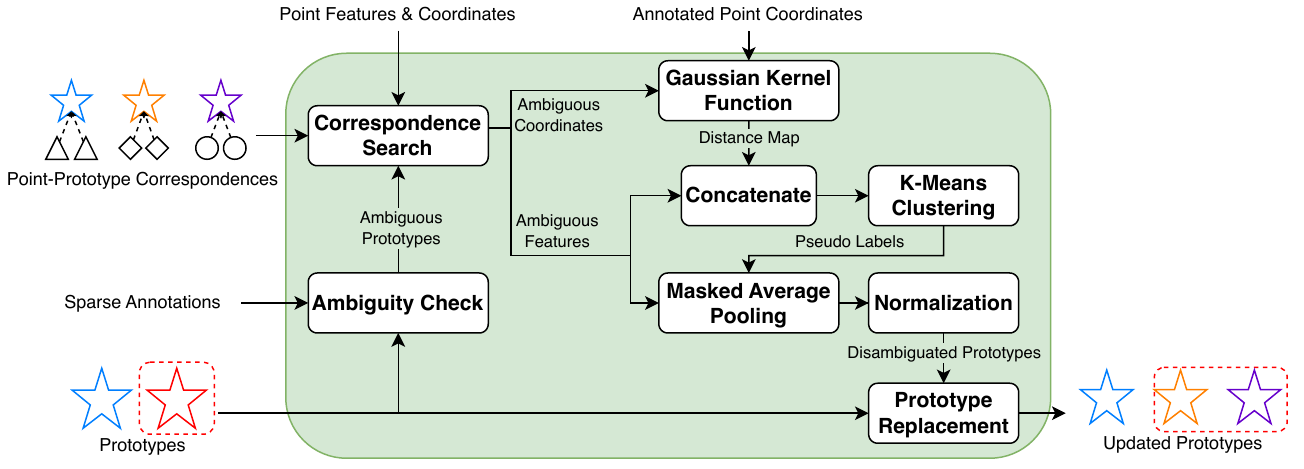}
\caption{
Workflow of prototype disambiguation.
Ambiguous prototypes are subsequently updated guided by sparse human annotations, ensuring class-specific discriminability of prototypes.}
\label{fig:method_prototype_disambiguation}
\end{figure*} 

\subsection{Prototype initialization} \label{method:prototype_initialization}
The prototypes are initialized upon the extracted features along with additional categorical features from the closed-world predictions, as demonstrated in \Cref{fig:method_prototype_initialization}.
We first augment the point features $\mathbf{F}$ based on the closed-world predictions $\text{Lin}(\mathbf{F})$, resulting in the category augmented features:
\begin{equation}
  \bar{\mathbf{F}}^c = \mathbf{F} \oplus \text{OE}(\text{Lin}(\mathbf{F})),
  \label{eq:category_augmented_features}
\end{equation}
where $\text{OE}(.)$ represents the one-hot encoding and $\oplus$ stands for the concatenation operation.
Next, we conduct K-Means\cite{lloyd-1982-tit} clustering upon the augmented features $\bar{\textbf{F}}^c$ to obtain the indication matrix $\mathbf{I} \in\{0,1\}^{n\times K}$: 
\begin{equation}
  \mathbf{I} = \text{KM}(\bar{\mathbf{F}}^c),
\end{equation}
where $\text{KM}(.)$ represents the K-Means clustering.
After that, we perform masked average pooling (MAP) on the original point features $\mathbf{F}$ based on the indication matrix $\mathbf{I}$ to construct the prototypes $\mathbf{Pt} = \{\mathbf{Pt}_k\}_{k=1}^{K}$, in which:
\begin{equation}
\begin{aligned}
  \mathbf{Pt}_k&=\text{Norm}(\mathop{\text{MAP}}\limits_{\mathbf{I}}(\mathbf{F})) \\
  &=\text{Norm}(\frac{\sum_{i = 1}^{N}\mathbf{F}_{i}\times \mathbf{I}_{i,k}}{\sum_{i = 1}^{N}\mathbf{I}_{i,k}}),
\end{aligned}
\end{equation}
where $\text{Norm(.)}$ represents the $l2$ normalization operation, and $\mathbf{I}_{i,k} \in \{0,1\}$ indicates the correspondences between the point $P_i$ and the $k$-th cluster.
Compared to directly clustering on the point features, the augmented features include more information from the pre-trained segmentor, resulting in more discriminative prototypes.

\begin{figure*}[!t]
\centering
\includegraphics[width=0.9\textwidth]{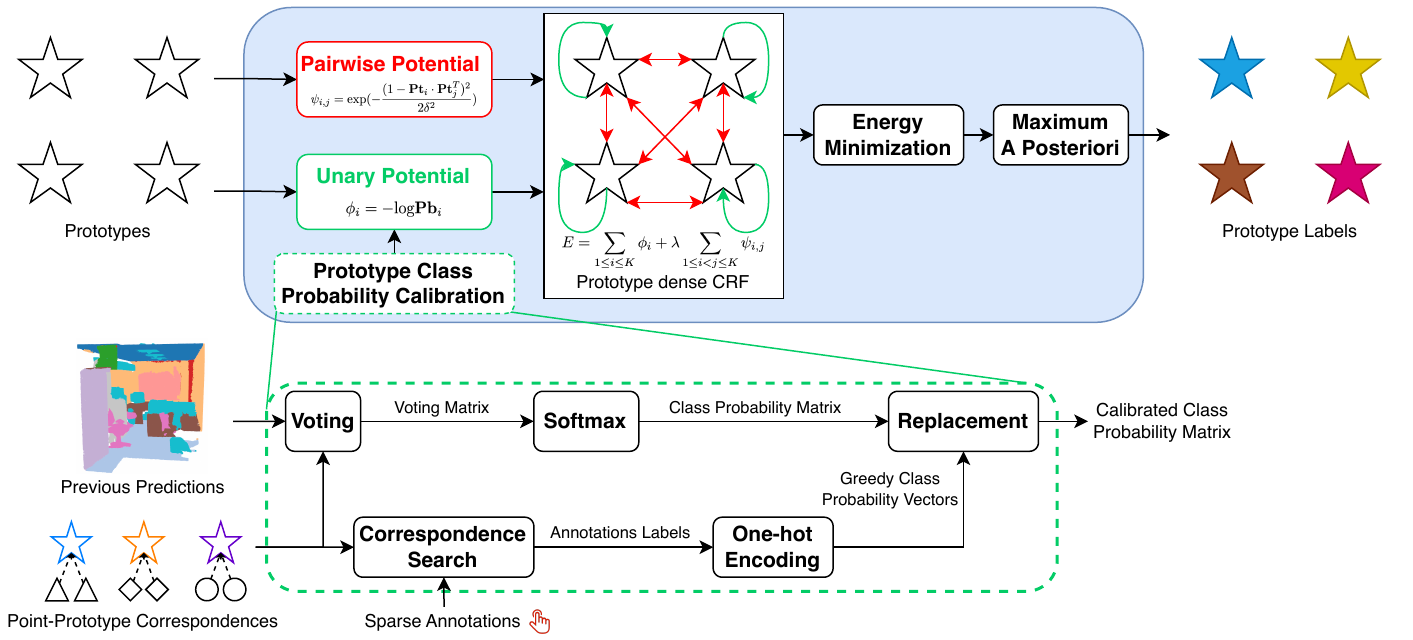}
\caption{
Workflow of prototype label assignment.
This module achieves context-aware label assignment via conditional random field optimization, integrating feature similarity with sparse human annotations to ensure semantically coherent prototype labeling.
}
\label{fig:method_prototype_label_assignment}
\end{figure*} 

\subsection{Prototype disambiguation} \label{method:prototype_disambiguation}
We propose the prototype disambiguation, allowing for less prototypes during initialization and using human sparse annotations to further refine the prototypes in a hierarchical way.
\Cref{fig:method_prototype_disambiguation} shows the workflow of prototype disambiguation.
A prototype will be regarded as an ambiguous one if its corresponding points contain annotations of different classes, where the correspondences are determined by the point-prototype correspondence matrix $\mathbf{CM} \in \mathbb{R}^{n\times K}$:
\begin{equation}
\label{eq:correspond_mat}
  \mathbf{CM} = \text{OE}(\text{Norm}(\mathbf{P}) \cdot \mathbf{Pt}^{T}).
\end{equation}
To handle the ambiguity, we propose the prototype disambiguation scheme by re-clustering upon the ambiguous points with auxiliary features, as illustrated in \Cref{fig:method_prototype_disambiguation}.
Specifically, for each ambiguous prototype, we build a distance map for each annotation, representing the Euclidean distance between the annotation and all ambiguous points to the prototype.
Given an ambiguous prototype $\mathbf{Pt}^a$ with $n_0$ ambiguous points $\mathbf{P}^a =\{\mathbf{P}^a_i\}_{i=1}^{n_0}$ and $n_1$ annotated points $\mathbf{Q}^a=\{\mathbf{Q}^a_j\}_{j=1}^{n_1} \in \mathbb{R}^{n_1\times d_0} \subset \mathbf{P}^a$, we have the distance map $\mathbf{D}^a\in \mathbb{R}^{n_0\times n_1}$, whose element is derived as:
\begin{equation}
  \mathbf{D}^a_{i,j} = \text{exp}(- \frac{\left\lVert \mathbf{P}^a_i - \mathbf{Q}^a_j \right\rVert_2}{2\sigma^2}),
\end{equation}
where $\sigma$ is a hyper-parameter, set to 1.0 by default.
We take the distance map as the auxiliary features to augment the original features, resulting in distance augmented features:
\begin{equation}
  \bar{\mathbf{F}}^{a} = \mathbf{F}^a\oplus \mathbf{D}^a,
  \label{eq:distance_augmented_features}
\end{equation}
where $\mathbf{F}^a$ represents the point features corresponding to the ambiguous prototype $\mathbf{Pt}^a$.
Based on the distance augmented features $\bar{\mathbf{F}}^{a}$, we construct the disambiguated prototypes $\mathbf{Pt}^{da}$ in a similar way to prototype initialization.
Specifically, we first derive the indication matrix $\mathbf{I}^a$ on those augmented features by K-Means clustering:
\begin{equation}
  \mathbf{I}^a = \text{KM}(\bar{\mathbf{F}}^{a}). 
\end{equation}
Then, we perform masked average pooling on the ambiguous point features based on the indication matrix, resulting in the disambiguated prototypes $\mathbf{Pt}^{da} = \{\mathbf{Pt}^{da}_k\}_{k=1}^{K^{da}}$, in which:
\begin{equation}
\begin{aligned}
  \mathbf{Pt}^{da}_k&=\text{Norm}(\mathop{\text{MAP}}\limits_{\mathbf{I}^a}(\mathbf{F}^a)) \\
  &=\text{Norm}(\frac{\sum_{i = 1}^{N}\mathbf{F}^a_{i}\times \mathbf{I}^a_{i,k}}{\sum_{i = 1}^{N}\mathbf{I}^a_{i,k}}).
\end{aligned}
\end{equation}
We set the number of disambiguated prototypes $K^{da}$ as the number of classes that appear in the annotations belonging to the ambiguous prototype.
Finally, the ambiguous prototypes are replaced with the disambiguated ones, resulting in the updated prototypes:
\begin{equation}
  \mathbf{Pt}^{t} = (\mathbf{Pt}^{t-1} \setminus \mathbf{Pt}^a) \cup \mathbf{Pt}^{da}.
\end{equation}

\subsection{Prototype label assignment} \label{method:prototype_label_assigment}
We assign prototype labels based on human sparse annotations and previous predictions.
\Cref{fig:method_prototype_label_assignment} shows the complete workflow of prototype label assignment, where a dense conditional random field (CRF) is build upon the prototypes for context-aware prototype label assignments.
In the prototype dense CRF, the unary potential is defined as the class probability of each prototype, which is calculated by voting on the previous point-wise predictions and further calibrated by human sparse annotations.
The pairwise potential is calculated on all possible prototype pairs by their cosine similarities.
We achieve the optimized class probabilities by solving the dense CRF, and finally achieve the prototype labels by maximum a posterior.

\subsubsection{Prototype class probability calibration} \label{method:prototype_class_probability_calibration}
We calculate the prototype class probabilities based on the previous predictions from the pre-trained closed-world segmentors or the former iteration for OW-Seg, and then calibrate them by human sparse annotations.
Specifically, we first compute the voting matrix $\mathbf{V}\in \mathbb{R}^{(|C_b|+1)\times K}$:
\begin{equation}
  \mathbf{V} = (\text{OE}(\text{Lin}(\mathbf{F})))^{T} \cdot \mathbf{CM},
\end{equation}
where $\mathbf{CM}$ is the point-prototype correspondence matrix mentioned in \Cref{eq:correspond_mat}.
Note that the novel classes are ignored in the open-set predictions and uniformly identified as the unknown class.
To achieve complete class probabilities, we compute the augmented voting matrix $\bar{\mathbf{V}} = [\bar{v}_{m}]_{(|C_b|+|C_n|+1)\times K}$:
\begin{equation}
  \bar{v}_{m} = 
\begin{cases}
v_{m}, & 1\leq m\leq |C_b|+1\\
v_{1}-1, & |C_b|+1 < m\leq |C_b|+|C_n|+1,
\end{cases}
\end{equation}
where $v_m$ and $\bar{v}_m$ represents the $m$-th row of the original and augmented voting matrixes respectively, $v_{1}$ corresponds to the unknown class, and $|C_n|$ denotes the number of possible novel classes, determined by user annotations. 
Finally, we achieve the previous prototype class probabilities with the softmax operation alongside the columns of the augmented voting matrix.

The proposed framework allows users to calibrate the class probabilities of prototypes.
Given a click with label, we first find its corresponding prototypes with the point-prototype correspondences.
After that, we convert the click label from a scalar to a greedy class probability vector with one-hot encoding.
We replace the previous class probability with the greedy one for the selected prototype, leading to class probability calibration.

\subsubsection{Prototype CRF construction}
We formulate a dense CRF upon the prototypes to infer optimized prototype label, in which the prototype class probabilities follow the Gibbs distribution:
\begin{equation}
  \mathbf{Pb}(\mathbf{Pl}|\mathbf{Pt}) = \frac{1}{Z(\mathbf{Pt})} \text{exp}(-\sum_{c \in \mathcal{C_G}} \Omega_c(\mathbf{Pl}_c|\mathbf{Pt})),
\end{equation}
where $Z(\mathbf{Pt})$ is a normalization constant, $\mathbf{Pl}$ represents the prototype label predictions, $\mathcal{C_G}$ represents a set of cliques in the graph $\mathcal{G=(V,E)}$ defined on the prototypes, and $\Omega_c(\mathbf{Pl}_c|\mathbf{Pt})$ indicates the potential for each clique $c$.
In the fully connected dense CRF, the Gibbs energy of labeling the prototypes consist of a unary potential term and a pairwise potential term:
\begin{equation}
  \mathbf{E} = \sum\limits_{1\leq i \leq K} \phi_i + \lambda \sum\limits_{1\leq i<j \leq K} \psi_{i,j},
\end{equation}
where $\lambda$ balances the effects between unary and pairwise potentials.

The unary potentials are defined as:
\begin{equation}
  \phi_i = - \text{log} \mathbf{Pb}_i,
\end{equation}
where the $\mathbf{Pb}_i$ represents the class probability of prototype $\mathbf{Pt}_i$ discussed in \Cref{method:prototype_class_probability_calibration}.
The pairwise potentials are defined as:
\begin{equation}
  \psi_{i,j} = \text{exp}(- \frac{(1 - \mathbf{Pt}_i \cdot \mathbf{Pt}_j^{T})^2}{2 \delta^2})
\end{equation}
where $\delta$ controls the degree of similarity.

\subsubsection{Inference}
We take the efficient approximate inference method by \cite{kruhenbuhl-2011-nips}.
Specifically, instead of directly calculating the $\mathbf{Pb}$, we take the mean field approximation to compute a distribution $\mathbf{Q}$ that optimizes the Gibbs energy while minimizing the KL-divergence $D_{KL}(\mathbf{Q}||\mathbf{Pb})$.
$\mathbf{Q}$ can be decomposed as a product of independent marginals:
\begin{equation}
  \mathbf{Q} = \prod_{i} \mathbf{Q}_i(\mathbf{y}_i),
\end{equation}
where i ranges from $1$ to $K$, and $\mathbf{y}_i$ represents the label of prototype $\mathbf{Pt}_i$.
We use $\mathbf{Q}_i$ to represent $\mathbf{Q}_i(\mathbf{y}_i)$ and use $\mathbf{Q}_i(l)$ to represent $\mathbf{Q}_i(\mathbf{y}_i=l)$ for simplicity.
The $\mathbf{Q}_i$ is iteratively approximated following \Cref{alg:crf_inference}.

\begin{algorithm}[t]
\caption{CRF inference by mean field approximation}
\label{alg:crf_inference}
\begin{algorithmic}[1]
\State $\mathbf{Q}_{i} \leftarrow \frac{1}{Z_{i}} \exp\{-\phi_{i}\}$
\Comment{Initialize $\mathbf{Q}$}
\While{not converged}
\State $\tilde{\mathbf{Q}}_{i}(l) \gets \sum\limits_{j\neq i} \psi_{i,j}\mathbf{Q}_{j}(l)$ for $l$ in $\mathcal{L}$ \Comment{Message passing}
\State $\hat{\mathbf{Q}}_{i} \gets \sum_{l\in\mathcal{L}} \tilde{\mathbf{Q}}_{i}(l)$ \Comment{Compatibility transform}
\State $\mathbf{Q}_{i} \gets \exp\{-\phi_{i} - \hat{\mathbf{Q}}_{i}\}$ \Comment{Local update}
\State normalize $\mathbf{Q}_{i}$
\EndWhile
\State $\mathbf{Q}_{i}^{*} \gets \mathbf{Q}_{i}$ \Comment{Output the optimized $\mathbf{Q}$}
\end{algorithmic}
\end{algorithm}

Finally, the prototype labels are obtained by maximum a posteriori:
\begin{equation}
\label{eq:prototype_label}
  \mathbf{Pl}_i = \mathop{\text{argmax}}\limits_{l\in\mathcal{L}} \mathbf{Q}_{i}^{*},
\end{equation}
where $\mathbf{Q}_{i}^{*}$ represents the optimized class probability by \Cref{alg:crf_inference}, $i$ ranges from 1 to $K$, and $\mathcal{L}$ ranges from 1 to $|\mathcal{C}_b|+|\mathcal{C}_n|+1$, covering all possible base and novel classes as well as the unknown class.

\begin{figure}[!t]
\centering
\includegraphics[width=0.48\textwidth]{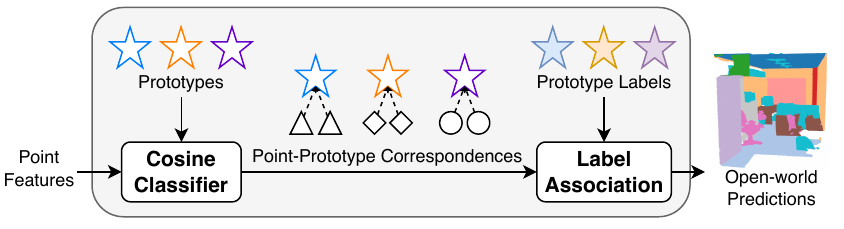}
\caption{
Workflow of prototype-based semantic segmentation.
This module achieves point-wise predictions based on point features and labeled prototypes.
}
\label{fig:method_prototype_semantic_segmentation}
\end{figure} 

\subsection{Prototype-based semantic segmentation}
We leverage the labeled prototypes to achieve the open-world predictions.
\Cref{fig:method_prototype_semantic_segmentation} shows the workflow of prototype-based semantic segmentation.
Given the constructed prototypes, we compute the point-prototype correspondence matrix $\mathbf{CM}$ as mentioned in \Cref{eq:correspond_mat}.
The point-prototype correspondences indicate the prototype belongings, i.e., the prototype that the points belong to.
With the prototype labels, we can obtain the open-world predictions $\mathbf{Y} \in \mathbb{R}^{n\times (|C_b|+|C_n|+1)}$ by associating labels from prototypes to their corresponding points:
\begin{equation}
  \mathbf{Y} = \mathbf{CM} \cdot \mathbf{Pl},
\end{equation}
where $\mathbf{Pl}$ is the prototype labels calculated in \Cref{eq:prototype_label}.

\begin{table*}[th]
    \small
    \centering
    \caption{The details of class partition of S3DIS and ScanNetv2}
    \begin{adjustbox}{width=0.9\textwidth}
    \begin{tabular}{p{1.5cm}|p{4cm}|p{2.5cm}|p{4.5cm}|p{2.5cm}}
        \toprule
        \multirow{2}{*}{Dataset} & \multicolumn{2}{c|}{Fold-0} & \multicolumn{2}{c}{Fold-1} \\
         & $\mathcal{C}_b$ & $\mathcal{C}_n$ & $\mathcal{C}_b$ & $\mathcal{C}_n$ \\
         \midrule
         S3DIS & beam, bookcase, ceiling, chair, door, floor, wall, clutter & table, window, column, board, sofa & beam, board, bookcase, ceiling, chair, column, door, floor, table, wall, clutter & window, sofa \\
         \hline
         ScanNetv2 & wall, floor, cabinet, bed, sofa, table, window, bookshelf, desk, curtain, chair, door, refrigerator, otherfurniture & sink, toilet, bathtub, shower curtain, picture, counter & wall, floor, cabinet, bed, sofa, table, window, bookshelf, picture, counter, desk, curtain, shower curtain, sink, bathtub, otherfurniture & chair, door, toilet, refrigerator \\
       \bottomrule
    \end{tabular}
    \end{adjustbox}
    \label{tab:class_partition}
\end{table*}

\section{Experiment} \label{sec:experiment}
\subsection{Experimental configuration} \label{subsec:exp_config}
\subsubsection{Datasets and setup}
We conduct experiments on the S3DIS \cite{armeni-2016-cvpr} and ScanNetv2 \cite{dai-2017-cvpr} datasets following the data processing strategy from \cite{wang-2019-tog,lai-2022-cvpr}.
Specifically, each point cloud scene is divided into non-overlapping $1m\times 1m$ blocks, where the raw points of each block is uniformly sampled by a grid of $0.02m$.

To adapt the data to OW-Seg, we split the semantic classes of each dataset into two non-overlapping subsets, representing the base and novel class sets.
We implement two class partition strategies to align with the setting of 3D GFS-Seg methods \cite{xu-2023-iccv} and IL-based OW-based methods \cite{xu-2024-cvpr}, corresponding to fold-0 and fold-1, respectively.
The details of class partition are demonstrated in \Cref{tab:class_partition}.

\subsubsection{Human simulation}
Due to the data scale and labor cost, we propose to simulate the human annotation behavior to efficiently and reproducibly evaluate the performance over the whole dataset.
Given a point cloud scene, as previously mentioned, we first divide it into several $1m\times 1m$ blocks for efficient processing.
Then, we sample a batch of adjacent blocks and stitch them into a sub-scene for annotation to simulate the real human field of view, where each sub-scene consists of up to 16 blocks in our experiments.
After that, we design four strategies to simulate real annotators in different scenarios, including the one-novel-class-one-click (ONCOC) strategy, one-class-one-click (OCOC) strategy, iterative strategy and iterative one-novel-class-one-click (IONCOC) strategy.

The OCOC strategy simulates an annotator who provides one click for each class occurring in the input, while the ONCOC strategy only focus on the novel classes.
To locate the click for each (novel) class, we cluster points of the same (novel) class by DBSCAN\cite{ester-1996-kdd} and pick the centroid point of the dominant cluster with the most points.
The iterative strategy simulates an annotator who tends to iteratively correct the most obvious mis-segmented area until the annotation budget is used out.
To locate the click in each iteration, we first extract the mis-segmented points by comparing the predictions with the ground truth.
Then, we divide the mis-segmented points into several regions by DBSCAN clustering.
Finally, we select the nearest point to the geometric centroid of the cluster to annotate.
The annotation will be used to update the predictions, which are further referred to generate new annotations.
The IONCOC strategy combines the two strategies above, simulating an annotator who first provides one click for each novel class, and then iteratively provides corrective clicks until the annotation budget is exhausted.

Note that the OCOC and ONCOC strategies provide clicks at one time, while the iterative and IONCOC strategies provide one click once the predictions are updated.
We set the click budget to 20 clicks by default in our experiments for iterative and IONCOC strategies.
The loop will end early if the prediction quality meets the requirement that the DBSCAN clustering algorithm cannot locate mis-segmented area for annotation, which means there are no obvious mis-segmented areas in the input.

\subsubsection{Implementation details}
We employ the Stratified Transformer (ST) \cite{lai-2022-cvpr} and Dynamic Graph Convolutional Neural Network (DGCNN) \cite{wang-2019-tog} as the segmentation backbones considering their effectiveness in capturing local features and global contexts.
For the ST, we constrain the number of points per block to 20480.
For the DGCNN, we constrain the number to 2048 due to the complexity of constructing $k$-NN graphs.
We train the ST for 100 epochs following the setting of \cite{lai-2022-cvpr} and train the DGCNN for 200 epochs with the Adam optimizer, where we set the initial learning rate to 1e-3 and decay it by half after every 50 epochs.
Only points belonging to the base classes are labeled and the rest points are deemed as the unknown class during the base class learning.

After the base class learning, we construct prototypes as discussed in \Cref{method:prototype_initialization} where the initial number of prototypes is set to 30.
In the prototype disambiguation procedure, we set the $\sigma$ to 0.5 to calculate the auxiliary distance map.
In the prototype label assignment module, we set the parameter $\lambda$ and $\delta$ to 1.0, and specify the number of iterations for mean field approximation as 10.

\subsubsection{Evaluation protocols}
We evaluate the overall performance using the mean class intersection-over-union (mIoU):
\begin{equation}
  \text{mIoU} = \frac{1}{|\mathcal{C}|} \sum\limits_{c\in \mathcal{C}} \frac{\text{TP}_c}{\text{TP}_c + \text{FP}_c + \text{FN}_c},
\end{equation}
where the $\mathcal{C}$ represents the set of all possible classes, $\text{TP}_c$ (True Positive), $\text{FP}_c$ (False Positive) and $\text{FN}_c$ (False Negative) are the number of points correctly predicted as class $c$, incorrectly predicted as class $c$, and belonging to class $c$ but incorrectly predicted as other classes, respectively.
We denote $\text{mIoU}_b$, $\text{mIoU}_n$ and $\text{mIoU}_a$ as the mIoU for base classes $\mathcal{C}_b$, novel classes $\mathcal{C}_n$ and all classes $\mathcal{C}_a = \mathcal{C}_b \cup \mathcal{C}_n$.

To alleviate the bias towards the base classes, following \cite{ye-2021-ijcv}, we employ the harmonic mean (HM) of $\text{mIoU}_b$ and $\text{mIoU}_n$ as the additional protocol to evaluate the overall performance, defined as:
\begin{equation}
  \text{HM} = \frac{2 \times \text{mIoU}_b \times \text{mIoU}_n}{\text{mIoU}_b + \text{mIoU}_n}.
\end{equation}

\begin{table*}[!th]
    \centering
    \small
    \caption{Comparison with GFS-Seg methods.
    HOW-Seg is evaluated under different annotation settings, including ONCOC (one-novel-class-one-click), OCOC (one-class-one-click), Iterative-$x$ ($x$ iterative clicks) and IONCOC-$x$ ($x$ iterative clicks starting from ONCOC).
    Optimal and sub-optimal results are highlighted in \textbf{bold} and \underline{underline}, respectively.}
    \begin{tabular}{p{2.2cm}p{1.6cm}p{1.7cm}|cccc|cccc}
        \toprule
        \multirow{2}{*}{Methods} & \multirow{2}{*}{Backbone} & \multirow{2}{*}{Setting} & \multicolumn{4}{c|}{S3DIS} & \multicolumn{4}{c}{ScanNetv2} \\
         & & & $\text{mIoU}_b$ & $\text{mIoU}_n$ & $\text{mIoU}_a$ & HM & $\text{mIoU}_b$ & $\text{mIoU}_n$ & $\text{mIoU}_a$ & HM \\
        \midrule
        \multirow{2}{*}{AttMPTI\cite{zhao-2021-cvpr}} & \multirow{2}{*}{DGCNN} & 1-shot & 21.89 & 11.39 & 17.05 & 14.95 & 12.97 & 1.62 & 9.57 & 2.88 \\
        & & 5-shot & 34.90 & 16.08 & 26.21 & 21.99 & 16.31 & 3.12 & 12.35 & 5.21 \\
        \hline
        \multirow{2}{*}{PIFS\cite{cermelli-2021-bmvc}} & \multirow{2}{*}{DGCNN} & 1-shot & 57.85 & 14.59 & 37.88 & 23.31 & 35.80 & 2.54 & 25.82 & 4.75 \\
        & & 5-shot & 56.99 & 19.66 & 39.76 & 29.23 & 35.14 & 3.21 & 25.56 & 5.88 \\
        \hline
        \multirow{2}{*}{CAPL\cite{tian-2022-cvpr}} & \multirow{2}{*}{DGCNN} & 1-shot & 72.80 & 23.87 & 50.22 & 35.67  & 38.70 & 10.59 & 30.27 & 16.53 \\
        & & 5-shot & 73.56 & 35.18 & 55.85 & 47.51  & 38.22 & 14.39 & 31.07 & 20.88  \\
        \hline
        \multirow{2}{*}{GFS-3DSeg\cite{xu-2023-iccv}} & \multirow{2}{*}{DGCNN} & 1-shot & 74.10 & 29.66 & 53.58 & 43.62 & 40.06 & 14.78 & 32.47 & 21.55  \\
        & & 5-shot & 73.61 & 43.26 & 59.60 & 54.42 & 40.18 & 18.58 & 33.70 & 25.39  \\
        \hline
        \multirow{2}{*}{PseudoEmbed\cite{tsai-2024-eccv}} & \multirow{2}{*}{DGCNN} & 1-shot & 74.54 & 39.78 & 58.50 & 51.34  & 40.47 & 15.57 & 33.00 & 22.47  \\
        & & 5-shot & 74.77 & 50.23 & 63.44 & 60.06 & 40.42 & 23.34 & 35.30 & 29.55  \\
        \hline
        \multirow{14}{*}{HOW-Seg (ours)} & & Baseline & 78.52 & 0 & 42.28 & 0 
        & 48.01 & 0 & 33.61 & 0 \\
       & & ONCOC & 73.90 & 51.06 & 63.36 & 60.39 
       & 47.99 & 28.71 & 42.20 & 35.92 \\
       &  & OCOC & 72.19 & 54.09 & 63.84 & 61.84 
       & 53.87 & 28.02 & 46.11 & 36.86 \\
       & DGCNN & Iterative-5 & 77.46 & 59.49 & 69.17 & 67.30 
       & 58.18 & 27.07 & 48.85 & 36.95 \\
       & (n\_blocks=4) & IONCOC-5 & 77.97 & 65.74 & 72.33 & 71.34 
       & 58.11 & 30.76 & 49.90 & 40.23  \\  
       & & Iterative-10 & 79.78 & 68.42 & 74.54 & 73.66 
       & 62.74 & 31.49 & 53.37 & 41.93 \\
       & & IONCOC-10 & 80.34 & 72.57 & 76.75 & 76.25 
       & 62.77 & 35.14 & 54.48 & 45.05 \\
        \cline{2-11}
         & & Baseline & 84.60 & 0 & 45.55 & 0 & 72.82 & 0 & 50.97 & 0 \\
       & & ONCOC & 83.79 & 67.06 & 76.07 & 74.50 & 60.18 & 50.68 & 57.33 & 55.02 \\
       & & OCOC & 85.74 & 66.66 & 76.93 & 75.00 & 65.11 & 49.01 & 60.28 & 55.92 \\
       & ST & Iterative-10 & 87.85 & 76.89 & 82.79 & 82.01 & 68.27 & 57.04 & 64.90 & 62.15 \\
       & (n\_blocks=16) & IONCOC-10 & 88.41 & 81.61 & 85.27 & 84.87 & 67.87 & 62.88 & 66.37 & \underline{65.28} \\
       & & Iterative-20 & 90.56 & 84.13 & \underline{87.59} & \underline{87.23} & 68.97 & 61.9 & \underline{66.85} & 65.24 \\
       & & IONCOC-20 & 90.53 & 85.47 & \textbf{88.20} & \textbf{87.93} & 69.15 & 65.65 & \textbf{68.1} & \textbf{67.36} \\  
        \bottomrule
    \end{tabular}
    \label{tab:gfsseg}
\end{table*}

\subsection{Comparison with GFS-Seg methods}
GFS-Seg is a special case of FS-Seg.
Similar to our setting, it predicts labels for both base and novel classes simultaneously.
The difference is GFS-Seg leverages densely annotated support data containing the novel classes present in the query data as the support information, while we rely on sparse annotations on the query data.
We compare our method with the SOTA GFS-Seg methods on S3DIS and ScanNetv2 datasets.
Among the comparative methods, attMPTI \cite{zhao-2021-cvpr} is the pioneering work in FS-Seg for point clouds, which is designed to predict labels of only novel classes during testing given support data.
To make it compatible to the setting of GFS-Seg, we follow the procedure of \cite{xu-2023-iccv} that preserves base prototypes before finishing training and merges them with novel prototypes generated by support data.
PIFS \cite{cermelli-2021-bmvc} fine-tunes on each novel session to incrementally update prototypes of the novel classes.
CAPL \cite{tian-2022-cvpr} is the seminal work of 2D GFS-Seg that predicts labels of both base and novel classes simultaneously.
GFS-3DSeg \cite{xu-2023-iccv} builds upon the framework of CAPL and introduces geometric words to facilitate GFS-Seg for 3D point clouds.
PseudoEmbed \cite{tsai-2024-eccv}, the current state-of-the-art 3D GFS-Seg method, employs foundation models and text embeddings to force clearer base/novel class differentiation.
Those GFS-Seg methods are evaluated under 1-shot and 5-shot settings.
We implement our HOW-Seg method with backbones of DGCNN for alignment with comparative methods and of ST for better segmentation performance.
As a human-in-the-loop method, we evaluate the proposed HOW-Seg with different human simulation strategies, including ONCOC, OCOC, Iterative and IONCOC, which have been discussed in \Cref{subsec:exp_config}.

\begin{figure*}[!t]
\centering
\includegraphics[width=1.\textwidth]{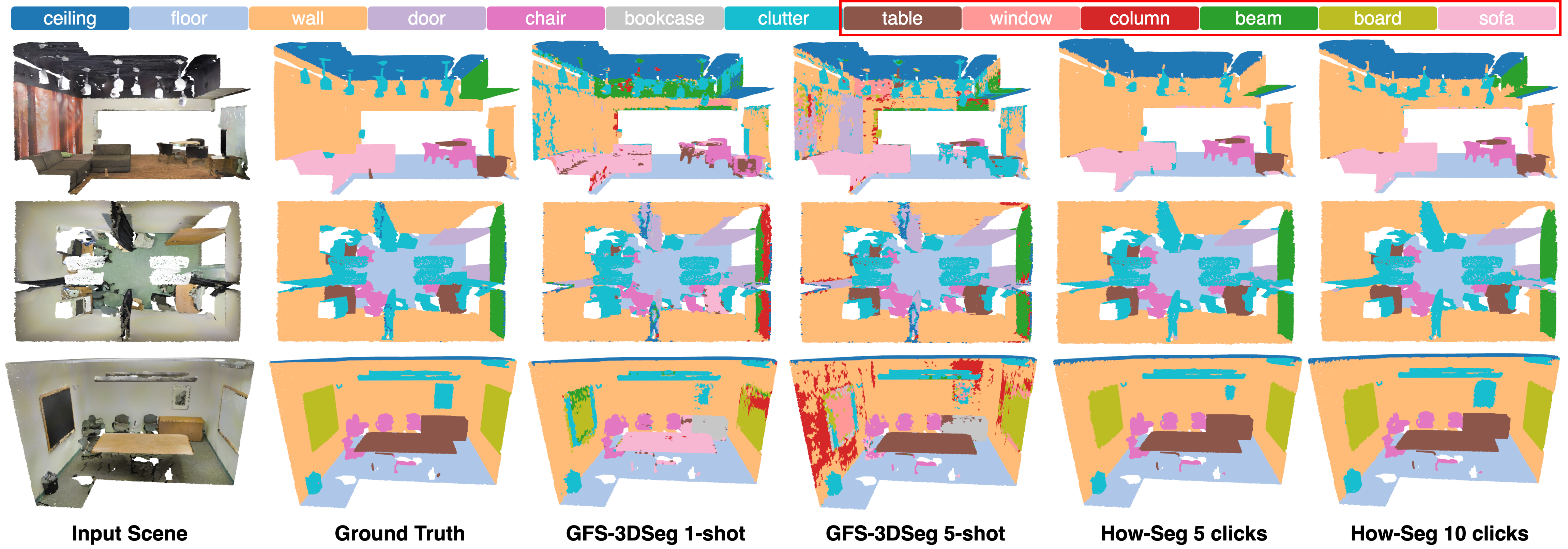}
\caption{
Qualitative results of the proposed HOW-Seg with 5 and 10 clicks on S3DIS in comparison to the GFS-3DSeg \cite{xu-2023-iccv} with 1-shot and 5-shot settings.
The novel classes are outlined with red rectangles.
}
\label{fig:exp_quali_s3dis}
\end{figure*}

\begin{figure*}[!t]
\centering
\includegraphics[width=1.\textwidth]{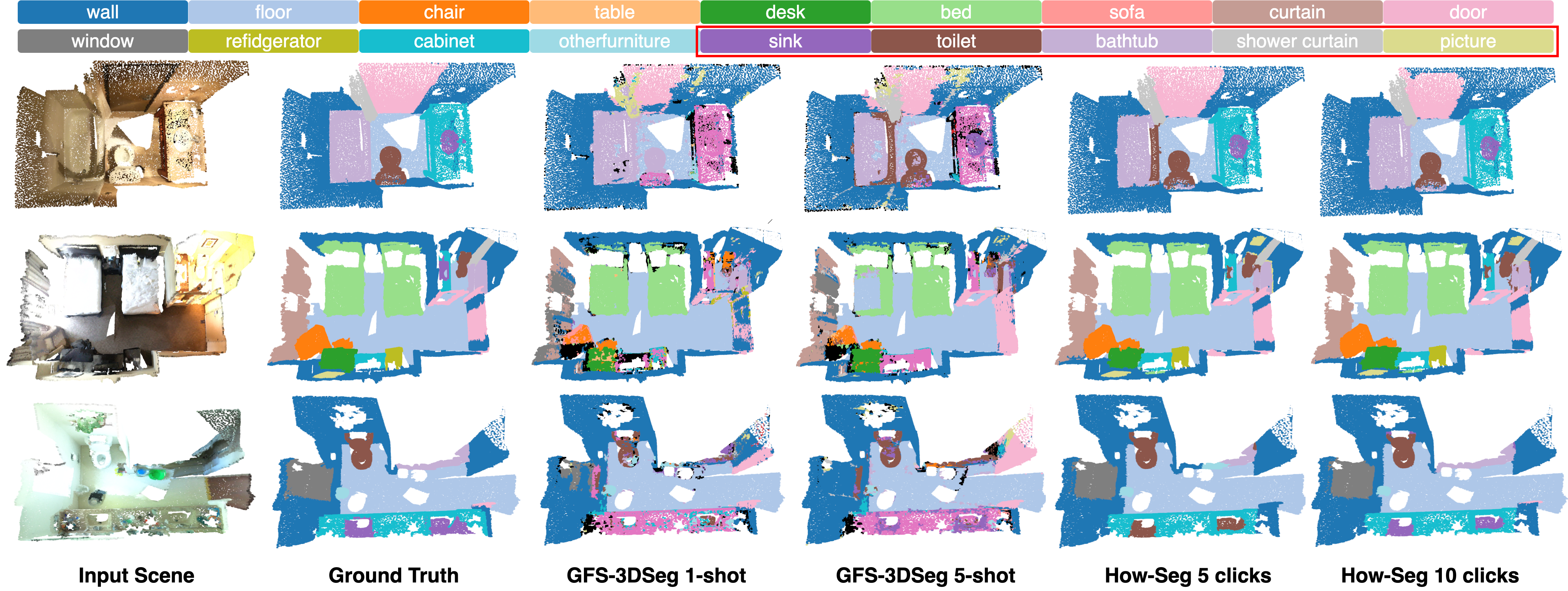}
\caption{
Qualitative results of the proposed HOW-Seg with 5 and 10 clicks on ScanNetv2 in comparison to the GFS-3DSeg \cite{xu-2023-iccv} with 1-shot and 5-shot settings.
The novel classes are outlined with red rectangles.
}
\label{fig:exp_quali_scannetv2}
\end{figure*}

As demonstrated in \Cref{tab:gfsseg}, HOW-Seg achieves significantly better results than the state-of-the-art GFS-Seg methods with a few clicks.
For example, HOW-Seg obtains mIoU of 88.2\% and HM of 87.93\% within 20 clicks per scene (consisting of 16 $1m\times 1m$ blocks), showing great segmentation capability for both base and novel classes.
In contrast, the current state-of-the-art GFS-Seg method, PseudoEmbed \cite{tsai-2024-eccv}, only achieves mIoU of 63.44\% and HM of 60.06\% under the 5-shot setting, which means 5 annotated samples are required for each novel class to support the segmentation process.
\Cref{fig:exp_quali_s3dis,fig:exp_quali_scannetv2} present more intuitive qualitative comparisons between GFS-Seg methods and ours.
Based on query-driven prototypes and real-time human feedback, HOW-Seg achieves results with high fidelity to the ground truth.
In contrast, the qualitative results of GFS-Seg exhibit noticeable prediction noises, where smooth and homogeneous regions are intermingled with irregularly distributed and inconsistent predictions.
Those artifacts stem from biased class prototypes extracted from pre-collected support samples, which induce misaligned decision boundaries and produce ambiguous predictions for the query sample.
Both of the quantitative and qualitative results demonstrate the superiority of HOW-Seg in producing high quality segmentation masks in open-world scenarios.

While HOW-Seg requires human intervention during each segmentation process, whereas GFS-Seg relies solely on pre-collected support samples, HOW-Seg achieves comparable performance to GFS-Seg's 5-shot setting with minimal human-in-the-loop annotations (e.g., one novel class one click), as shown in \Cref{tab:gfsseg}.
By incrementally increasing the click budget, HOW-Seg can progressively refine its outputs to near-ground-truth segmentation quality.
In contrast, the segmentation quality of GFS-Seg dose not scale directly with the number of support samples.
As demonstrated in \Cref{fig:exp_quali_s3dis,fig:exp_quali_scannetv2}, HOW-Seg achieves segmentation results nearly indistinguishable from the ground truth with just 5 clicks, while GFS-Seg exhibits pronounced segmentation errors (e.g., mis-calibrated boundaries, misclassified and fragmented regions) under the 1-shot and 5-shot settings.
Moreover, post-processing those erroneous predictions demands significantly higher human labor.

\begin{table*}[!th]
    \centering
    \small
    \caption{Comparison with IL-based OW-Seg methods.
    HOW-Seg is evaluated under different annotation settings, including ONCOC (one-novel-class-one-click), OCOC (one-class-one-click), Iterative-$x$ ($x$ iterative clicks) and IONCOC-$x$ ($x$ iterative clicks starting from ONCOC).
    Optimal and sub-optimal results are highlighted in \textbf{bold} and \underline{underline}, respectively.}
    \begin{tabular}{p{2.2cm}p{1.6cm}p{1.8cm}|cccc|cccc}
        \toprule
        \multirow{2}{*}{Methods} & \multirow{2}{*}{Backbone} & \multirow{2}{*}{Setting} & \multicolumn{4}{c|}{S3DIS} &  \multicolumn{4}{c}{ScanNetv2} \\
         &  &  & $\text{mIoU}_b$ & $\text{mIoU}_n$ & $\text{mIoU}_a$ & HM  & $\text{mIoU}_b$ & $\text{mIoU}_n$ & $\text{mIoU}_a$ & HM \\
        \midrule
        LwF\cite{li-2017-tpami} & PT & IL & 69.2 & 23.9 & 62.3 & 35.53 & 50.9 & 12.0 & 43.1 & 18.77 \\
        \cline{1-11}
        REAL\cite{cen-2022-eccv} & PT & IL & 70.3 & 61.3 & 68.9 & 65.49  & 71.6 & 67.3 & \underline{70.8} & 69.01 \\
        \cline{1-11}
        PDF\cite{xu-2024-cvpr} & PT & IL & 70.3 & 64.3 & 69.4 & 67.17  & 71.7 & 68.3 & \textbf{71.0} & \underline{69.62}  \\
        \hline        
       \multirow{14}{*}{HOW-Seg (ours)} & & Baseline & 71.64 & 0 & 60.62 & 0 & 46.90 & 0 & 37.52 & 0 \\        
       & & ONCOC & 64.92 & 68.59 & 65.49 & 66.71 
       & 46.62 & 32.88 & 43.87 & 38.56 \\
       & & OCOC & 64.78 & 67.19 & 65.15 & 65.96 
       & 52.73 & 34.48 & 49.08 & 41.70 \\
       & DGCNN & Iterative-5 & 74.54 & 76.80 & 74.89 & 75.66 
       & 54.20 & 40.22 & 51.41 & 46.18 \\
       & (n\_blocks=4) & IONCOC-5 & 74.66 & 76.49 & 74.94 & 75.56 
       & 54.23 & 40.69 & 51.52 & 46.50  \\
       & & Iterative-10 & 77.96 & 80.13 & 78.29 & 79.03 
       & 58.21 & 43.95 & 55.36 & 50.09 \\
       & & IONCOC-10 & 77.95 & 78.77 & 78.08 & 78.36 
       & 58.45 & 45.44 & 55.85 & 51.13 \\
        \cline{2-11}
         & & Baseline & 81.63 & 0 & 69.07 & 0 & 69.73 & 0 & 55.78 & 0 \\
       & & ONCOC & 80.26 & 77.63 & 79.86 & 78.92 
       & 57.83 & 59.93 & 58.25 & 58.86 \\
       & & OCOC & 82.06 & 81.71 & 82.00 & 81.89 
       & 61.27 & 65.95 & 62.21 & 63.52 \\
       & ST & Iterative-10 & 87.62 & 87.28 & 87.57 & 87.45  
       & 64.73 & 66.27 & 65.04 & 65.49 \\
       & (n\_blocks=16) & IONCOC-10 & 87.15 & 86.69 & 87.08 & 86.92 
       & 65.33 & 72.58 & 66.78 & 68.77 \\
       & & Iterative-20 & 89.66 & 91.86 & \textbf{90.00} & \underline{90.75}  
       & 65.93 & 68.02 & 66.35 & 66.96 \\
       & & IONCOC-20 & 89.58 & 92.06 & \underline{89.96} & \textbf{90.80} 
       & 65.4 & 74.84 & 67.35 & \textbf{69.85} \\
        \bottomrule
    \end{tabular}
    \label{tab:owseg}
\end{table*}

\subsection{Comparison with IL-based OW-Seg methods}
IL-based OW-Seg methods allow models to predict labels for novel classes by introducing an offline incremental learning stage, where additional supervision signals for these novel classes are needed.
In contrast, our method operates in fully test-time after base class learning and learns to predict novel class labels through real-time feedback from human within the segmentation loop.
In addition, we treat the novel classes as the unknown class during base class learning following the setting of \cite{xu-2023-iccv} and \cite{chakravarthy-2024-ijcv}, while the IL-based OW-Seg methods typically ignore the novel classes, leading to difference in base class learning.
Despite the setting differences, we conduct a comparative analysis with SOTA IL-based 3D OW-Seg methods for reference, including REAL \cite{cen-2022-eccv} and PDF \cite{xu-2024-cvpr}.
REAL is the pioneering work in IL-based OW-Seg for outdoor LiDAR point clouds.
PDF is the current SOTA method applied to indoor scene point clouds.
We also include LwF \cite{li-2017-tpami} as the baseline method of incremental learning.

\Cref{tab:owseg} shows the performance comparison between our method and existing IL-based OW-Seg methods with results reproduced from \cite{xu-2024-cvpr}.
The proposed HOW-Seg demonstrates comparable performance to the SOTA IL-based OW-Seg method in the ONCOC setting and significantly outperforms it as more clicks are provided, highlighting its effectiveness
Though the IL-based OW-Seg methods experience novel class incremental learning under sufficient supervision, two key challenges remain: 
first, they rely on offline training samples to extract novel class knowledge, which results in suboptimal test-time performance due to the intra-class variance;
second, the novel class incremental learning process inevitably impacts the base class performance due to the catastrophic forgetting.
In contrast, our HOW-Seg learns to segment novel classes by leveraging instant feedback from test samples, thereby avoiding intra-class variance. 
Additionally, our method maintains fixed network parameters during testing, effectively alleviating catastrophic forgetting. 
As a human-in-the-loop framework, our method constructs a closed-loop segmentation system during testing that continuously refines predictions using human-provided feedback, leading to progressively better outcomes.

\subsection{Ablation studies}

\begin{table*}[!th]
    \centering
    \small
    \caption{Effects of click budget.
    We report the actual average number of clicks per scene, which is typically lower than the click budget due to early termination when the segmentation quality meets the predefined criteria.}
    \begin{tabular}{p{1.cm}|cccccc|cccccc}
        \toprule
        \multirow{2}{*}{Budget} & \multicolumn{6}{c|}{S3DIS} & \multicolumn{6}{c}{ScanNetv2} \\
        & $\text{mIoU}_b$ & $\text{mIoU}_n$ & $\text{mIoU}_a$ & HM & $\Delta(\text{mIoU}_a)$ & Clicks & $\text{mIoU}_b$ & $\text{mIoU}_n$ & $\text{mIoU}_a$ & HM & $\Delta(\text{mIoU}_a)$ & Clicks \\
        \midrule
        0 & 81.63 & 0 & 69.07 & 0 & - & 0 
        & 69.73 & 0 & 55.78 & 0 & - & 0 \\
        5 & 87.44 & 74.60 & 81.51 & 80.51 & 18.01\% & 4.75 
        & 66.59 & 58.07 & 64.03 & 62.04 & 14.79\% & 4.63 \\
        10 & 88.41 & 81.61 & 85.27 & 84.87 & 23.45\% & 9.82 
        & 67.87 & 62.88 & 66.37 & 65.28 & 18.99\% & 8.87 \\
        20 & 90.53 & 85.47 & 88.20 & 87.93 & 27.70\% & 19.45 
        & 69.15 & 65.65 & 68.10 & 67.36 & 22.09\% & 16.75 \\
        30 & 91.94 & 87.48 & 89.88 & 89.65 & 30.13\% & 27.69 
        & 70.03 & 64.92 & 68.50 & 67.38 & 22.80\% & 23.58 \\
        \bottomrule
    \end{tabular}
    \label{tab:ab_click}
\end{table*}

\subsubsection{Effects of click budget}
We investigate the segmentation performance under different click budgets.
As demonstrated in \Cref{tab:ab_click}, the segmentation performance generally improves as the click budget increases.
Note that the improvement rate gradually slows down as the click budget grows since the additional clicks provide diminishing returns in terms of information gain.
It is also noteworthy that the actual number of clicks is always less than the budget. 
This is because the human-in-the-loop segmentation process can be terminated early if the segmentation result has already met the required quality, even if the click budget has not been fully used. 
This early termination imitates the behavior of real annotators, helping to save unnecessary time and computations.
Considering the trade-off between the segmentation accuracy and annotation cost, we set the click budget to 20 in the subsequent experiments.

\subsubsection{Effects of prototype number} \label{exp:ab_protonum}
We investigate the effects of the initial prototype number on the overall performance.
As shown in \Cref{tab:ab_protonum}, when the prototype number is relatively small (e.g., 10), the segmentation performance of both base and novel classes is relatively poor. 
This is because the limited number of prototypes cannot provide sufficient discriminative ability to distinguish between different classes effectively.
As the prototype number increases, the performance improves significantly. 
For example, increasing the prototype number from 10 to 30 improves the mIoU on S3DIS from 85.71\% to 88.20\%, while only incurring marginal time cost less than 0.2s.
This is because more prototypes can better capture the diversity and complexity of the data, leading to better segmentation results.
However, as the number of prototypes further increases (e.g., to 70), the performance improves only marginally or even deteriorates, while the computational time rises substantially.
For instance, the mIoU on S3DIS of 70 prototypes is on-par that of 30 prototypes (i.e., 88.20\% v.s. 88.44\%), while incurring more than five times the computational cost (i.e., from 0.4071s to 2.28s).
This is likely due to the increased difficulty in assigning correct labels to a larger number of prototypes, which introduces extra computational burden in the prototype initialization process and noise in the inference process.
Through comprehensive experiments shown in \Cref{tab:ab_protonum}, we find that the performance almost reaches its peak with 30 initial prototypes on S3DIS and 50 on ScanNetv2, achieving a good balance between segmentation accuracy and computational efficiency.
In the following experiments, we set the number of initial prototypes to 30 and 50 on S3DIS and ScanNetv2, respectively.

\begin{table*}[!t]
    \centering
    \small
    \caption{Effects of prototype number (ProtoNum).}
    \begin{tabular}{p{1.2cm}|ccccc|ccccc}
        \toprule
        \multirow{2}{*}{ProtoNum} & \multicolumn{5}{c|}{S3DIS} & \multicolumn{5}{c}{ScanNetv2} \\
        & $\text{mIoU}_b$ & $\text{mIoU}_n$ & $\text{mIoU}_a$ & HM & Time(s) & $\text{mIoU}_b$ & $\text{mIoU}_n$ & $\text{mIoU}_a$ & HM & Time(s) \\
        \midrule
        10 & 88.92 & 81.96 & 85.71 & 85.30 & 0.2280
        & 66.76 & 62.35 & 65.44 & 64.48 & 0.1512 \\
        30 & 90.53 & 85.47 & \underline{88.20} & \underline{87.93} & 0.4071 
        & 69.37 & 63.69 & 67.67 & 66.41 & 0.2992 \\
        50 & 90.76 & 84.75 & 87.99 & 87.65 & 0.9441
        & 69.15 & 65.65 & \textbf{68.10} & \underline{67.36} & 0.7321 \\
        70 & 91.35 & 85.04 & \textbf{88.44} & \textbf{88.08} & 2.2800
        & 68.77 & 66.11 & \underline{67.97} & \textbf{67.41} & 1.9409 \\
        \bottomrule
    \end{tabular}
    \label{tab:ab_protonum}
\end{table*}

\begin{table*}[t]
    \centering
    \small
    \caption{Effects of prototype disambiguation (ProtoDis) under different prototype numbers.}
    \begin{tabular}{p{1.2cm}p{1.1cm}|ccccc|ccccc}
        \toprule
        \multirow{2}{*}{ProtoNum} & \multirow{2}{*}{ProtoDis} &  \multicolumn{5}{c|}{S3DIS} & \multicolumn{5}{c}{ScanNetv2} \\
        & & $\text{mIoU}_b$ & $\text{mIoU}_n$ & $\text{mIoU}_a$ & HM & $\Delta(\text{mIoU}_a)$ & $\text{mIoU}_b$ & $\text{mIoU}_n$ & $\text{mIoU}_a$ & HM & $\Delta(\text{mIoU}_a)$ \\
        \midrule
        \multirow{2}{*}{10} & \ding{55} & 76.92 & 55.44 & 67.00 & 64.44 & \multirow{2}{*}{27.93\%} & 58.36 & 29.82 & 49.80 & 39.47 & \multirow{2}{*}{31.41\%} \\
         & \checkmark & 88.92 & 81.96 & 85.71 & 85.30 & & 66.76 & 62.35 & 65.44 & 64.48 &  \\
        \hline
         \multirow{2}{*}{30} & \ding{55} & 84.41 & 70.45 & 77.97 & 76.80 & \multirow{2}{*}{13.12\%} & 64.46 & 52.09 & 60.75 & 57.62 & \multirow{2}{*}{11.39\%} \\
         & \checkmark & 90.53 & 85.47 & 88.20  & 87.93 & & 69.37 & 63.69 & 67.67 & 66.41 & \\
        \hline
         \multirow{2}{*}{50} & \ding{55} & 88.14 & 77.73 & 83.34 & 82.61 & \multirow{2}{*}{5.58\%} & 66.98 & 59.04 & 64.60 & 62.76 & \multirow{2}{*}{5.42\%} \\
         & \checkmark & 90.76 & 84.75 & 87.99 & 87.65 &  & 69.15 & 65.65 & 68.10 & 67.36 & \\
        \hline
         \multirow{2}{*}{70} & \ding{55} & 89.07 & 80.82 & 85.26 & 84.74 & \multirow{2}{*}{3.73\%} & 67.21 & 61.43 & 65.48 & 64.19 & \multirow{2}{*}{3.80\%} \\
         & \checkmark & 91.35 & 85.04 & 88.44 & 88.08 &  & 68.77 & 66.11 & 67.97 & 67.41 & \\
        \bottomrule
    \end{tabular}
    \label{tab:ab_protodis}
\end{table*}

\begin{figure*}[!t]
\centering
\includegraphics[width=1.\textwidth]{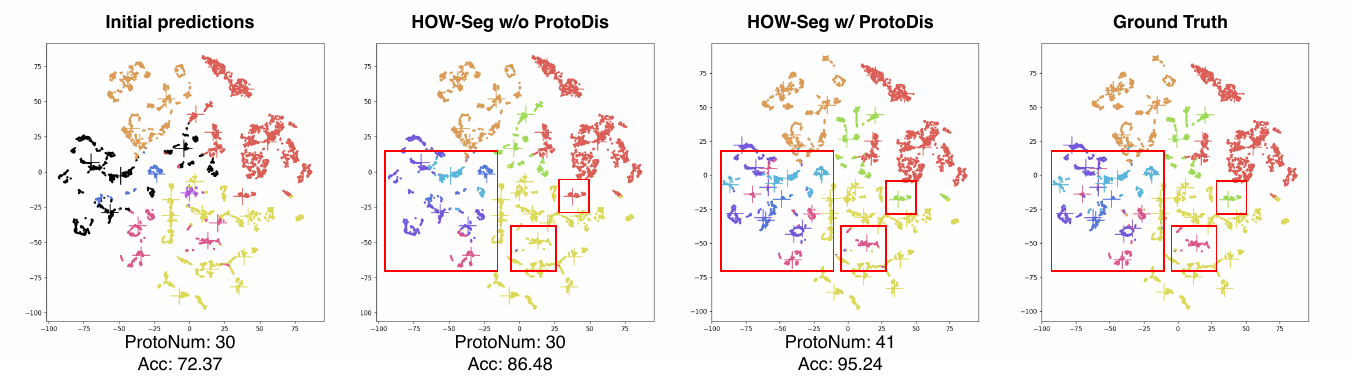}
\caption{
t-SNE visualization of last-layer features of an exemplar from the S3DIS, colored (from left to right) by: initial predictions, predictions of HOW-Seg without and with the prototype disambiguation (ProtoDis), and the Ground Truth.
We mark the prototypes with cross symbols and outline areas with significant differences using red rectangles. 
The instant number of prototypes (ProtoNum) and the corresponding point-wise classification accuracy (Acc) are presented below each prediction.
}
\label{fig:exp_tsne}
\end{figure*}

\begin{table*}[!t]
    \centering
    \small
    \caption{Effects of prototype label assignment (PLA) with prototype disambiguation (ProtoDis).
    We employ the IONCOC simulation strategy with click budget of 20.}
    \begin{tabular}{p{1cm}p{0.8cm}|ccccc|ccccc}
        \toprule
        \multirow{2}{*}{ProtoDis} & \multirow{2}{*}{PLA} & \multicolumn{5}{c|}{S3DIS} & \multicolumn{5}{c}{ScanNetv2} \\
        & & $\text{mIoU}_b$ & $\text{mIoU}_n$ & $\text{mIoU}_a$ & HM & $\Delta(\text{mIoU}_a)$ & $\text{mIoU}_b$ & $\text{mIoU}_n$ & $\text{mIoU}_a$ & HM & $\Delta(\text{mIoU}_a)$ \\
        \midrule
       \ding{55} & \ding{55} & 87.99 & 71.58 & 80.42 & 78.94 & - & 66.17 & 56.54 & 63.28 & 60.98 & - \\
        \ding{55} & \checkmark & 84.41 & 70.60 & 78.04 & 76.89 & -2.96\% & 66.98 & 59.04 & 64.60 & 62.76 & 2.09\% \\
        \checkmark & \ding{55} & 90.90 & 84.57 & \underline{87.98} & \underline{87.62} & 9.40\% & 68.55 & 66.27 & \underline{67.86} & \textbf{67.39} & 7.24\% \\
        \checkmark & \checkmark & 90.53 & 85.47 & \textbf{88.20} & \textbf{87.93} & 9.67\% & 69.15 & 65.65 & \textbf{68.10} & \underline{67.36} & 7.62\% \\
        \bottomrule
    \end{tabular}
    \label{tab:ab_protodis_pla}
\end{table*}

\subsubsection{Effects of prototype disambiguation}
Here we investigate the effects of prototype disambiguation (ProtoDis) in the contexts of different initial prototype numbers.
As shown in \Cref{tab:ab_protodis}, the ProtoDis generally improves the segmentation performance across different initial prototype numbers. 
The model discriminability is limited when the prototypes are sparse (e.g., 10). 
In this case, the ProtoDis breaks the original decision boundaries by hierarchically dividing ambiguous prototypes into discriminative sub-prototypes, leading to the most obvious performance increase, e.g., 27.93\% growth of mIoU on S3DIS and 31.41\% on ScanNetv2.
As the prototype number grows, the discriminability improves accordingly while increasing the computational cost, as discussed in \Cref{exp:ab_protonum}.
The introduction of ProtoDis not only enhances overall segmentation performance but, more importantly, enables the model to achieve comparable segmentation accuracy with fewer prototypes, which remarkably improves the model's efficiency.
As demonstrated in \Cref{tab:ab_protodis}, by using the ProtoDis, the mIoU reaches the peak with 30 prototypes on S3DIS and 50 prototypes on ScanNetv2.
To visually illustrate the effect of prototype disambiguation, we conduct the t-SNE analysis to show the feature and prototype distributions of HOW-Seg with and without the ProtoDis, as shown in \Cref{fig:exp_tsne}.
During the human-in-the-loop segmentation process, the initial 30 prototypes are decomposed into 41 prototypes by the ProtoDis, boosting the point-wise accuracy from 72.37\% to 95.24\%, significantly outperforming the non-ProtoDis group’s 86.48\%.

\subsubsection{Effects of prototype label assignment}
Here we jointly investigate the effects of prototype label assignment (PLA) and ProtoDis.
We design a variant that obtains the prototype labels by directly maximizing the initial prototype class probability instead of the proposed PLA module.
This variant serves as a baseline to compare when evaluating the impact of PLA.
As shown in \Cref{tab:ab_protodis_pla}, the combination of PLA and ProtoDis achieves the best overall segmentation performance.
It is noteworthy that when only PLA is applied (without ProtoDis), the overall performance actually decreases. 
This degradation is likely due to incorrect initial prototypes, which introduce confusion during the prototype label assignment and the final inference phases.
In contrast, when we apply ProtoDis, those ambiguous prototypes are decomposed into several more certain prototypes. 
This decomposition allows for clearer decision boundaries and more accurate label propagation through the constructed prototype CRF. 
As a result, the labels can be correctly propagated and assigned, leading to improved segmentation performance.

\section{Conclusion}
In this paper, we propose HOW-Seg, a human-in-the-loop framework for open-world point cloud semantic segmentation.
By constructing query-driven prototypes, our framework mitigates distribution shifts across different samples, resulting in unbiased class representations.
Through introducing the interactive prototype disambiguation and CRF-based prototype label assignment mechanisms, we obtain discriminative dynamic prototypes with context-aware labels, enabling high quality open-world semantic segmentation.
Crucially, HOW-Seg operates in a fully test-time manner without offline fine-tuning or additional support data, making it highly practical for real-world applications.
Extensive experiments on S3DIS and ScanNetv2 demonstrate HOW-Seg’s superiority over existing alternatives.

\ifCLASSOPTIONcompsoc
  \section*{Acknowledgments}
\else
  \section*{Acknowledgment}
\fi

This work was supported by the National Key Research and Development Program of China under Grant 2022QY0102. The authors would like to acknowledge the helpful comments and kindly suggestions provided by anonymous referees.

\bibliography{my_reference}

\vspace{-10mm}

\begin{IEEEbiography}[{\includegraphics[width=1in,height=1.25in,clip,keepaspectratio]
{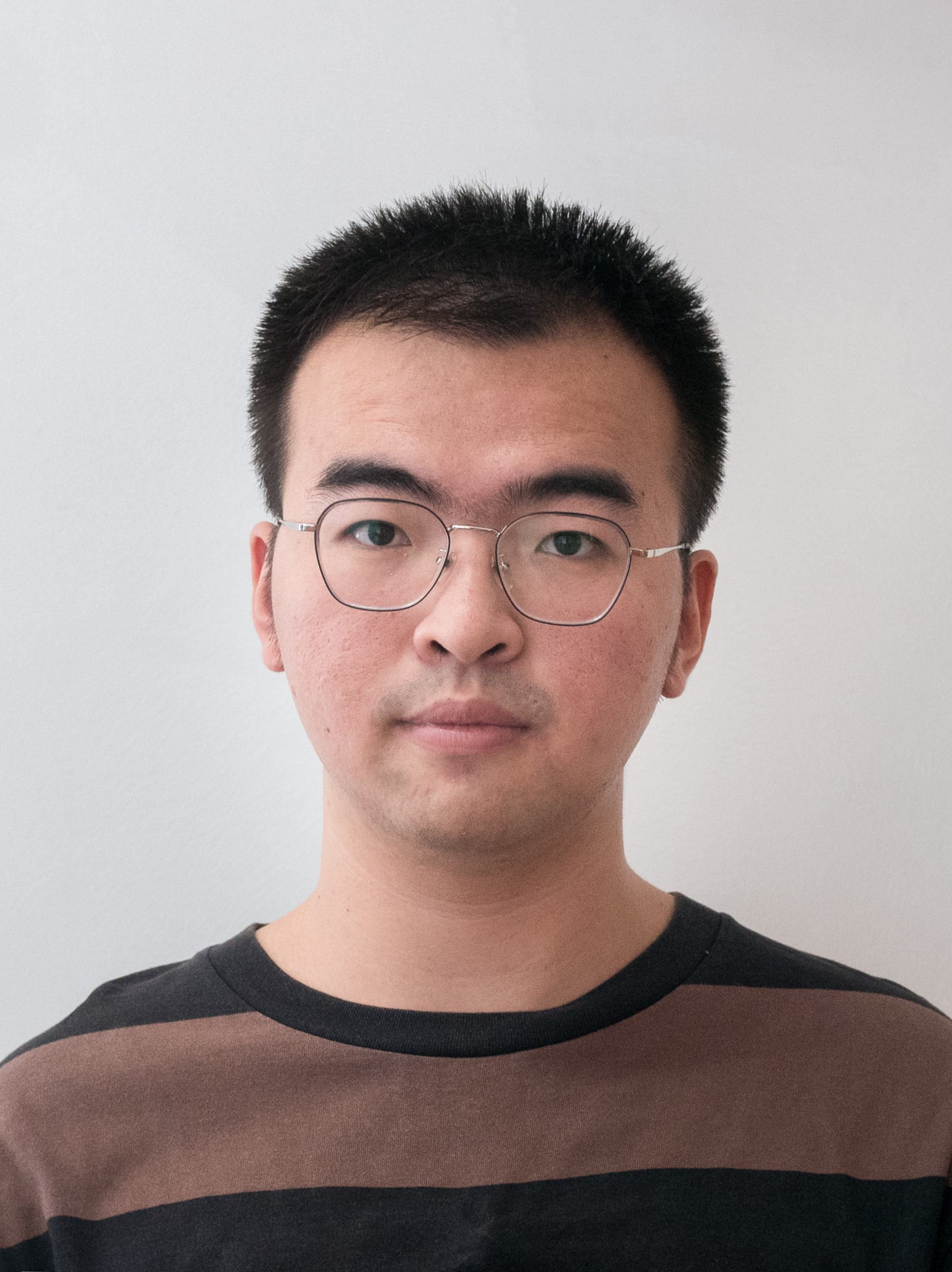}}]{Peng Zhang}
is currently pursuing the Ph.D. degree at the School of Automation, Nanjing University of Science and Technology, China. He received his B.S. degree from the College of Electrical Engineering, Henan University of Technology, in 2020. His research interests include computer graphics, 3D computer vision and point cloud processing.
\end{IEEEbiography}

\vspace{-10mm}
\begin{IEEEbiography}[{\includegraphics[width=1in,height=1.25in,clip,keepaspectratio]
{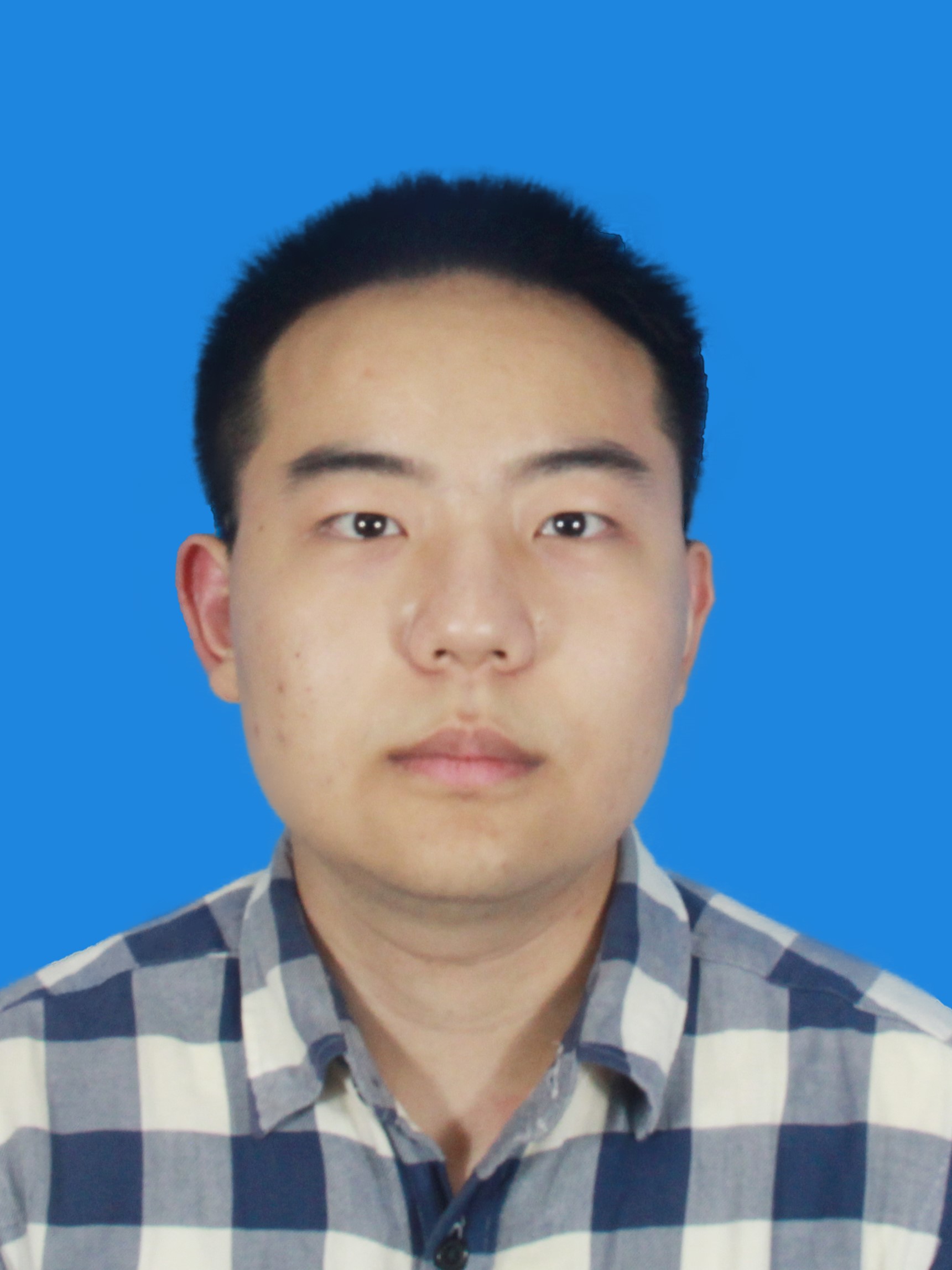}}]{Songru Yang}
is currently pursuing her M.S. degree at School of Automation, Nanjing University of Science and Technology.
He received the B.S. degree from School of Electrical Engineering and Information, Southwest Petroleum University, in 2024. 
His research interests are point cloud sampling and segmentation.
\end{IEEEbiography}

\vspace{-10mm}
\begin{IEEEbiography}[{\includegraphics[width=1in,height=1.25in,clip,keepaspectratio]
{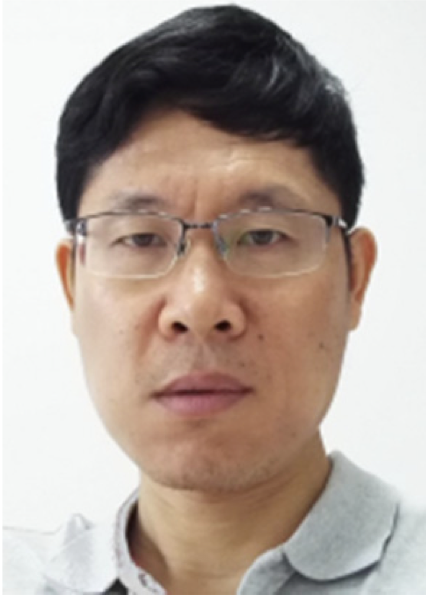}}]{Jinsheng Sun}
has been with the School of Automation, Nanjing University of Science and Technology (NUST), Nanjing, China, where he is currently a professor, since 1995. He received his B.S., M.S. and PhD of Control Science and Engineering from NUST in 1990, 1992 and 1995. From 2007 to 2009, he visited University of Melbourne as a Research Fellow at the Department of Electrical and Electronic Engineering. And in 2011, he was with the City University of Hong Kong as a senior Research Fellow. His research activity includes network congestion control, quality control and distributed control of multi-agent system.
\end{IEEEbiography}

\vspace{-10mm}
\begin{IEEEbiography}[{\includegraphics[width=1in,height=1.25in,clip,keepaspectratio]
{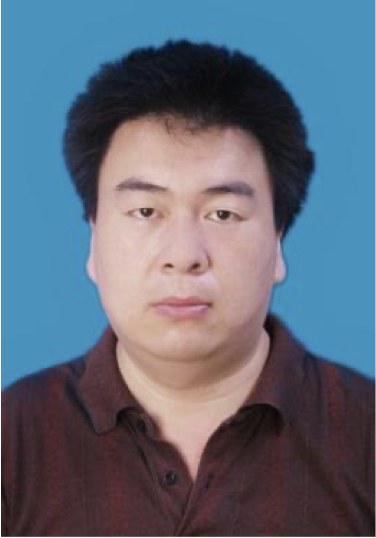}}]{Weiqing Li}
is currently an associate professor at the School of Computer Science and Engineering, Nanjing University of Science and Technology, China. He received the B.S. and Ph.D. degrees from the School of Computer Sciences and Engineering, Nanjing University of Science and Technology in 1997 and 2007, respectively. His current interests include computer graphics and virtual reality.
\end{IEEEbiography}


\vspace{-10mm}
\begin{IEEEbiography}[{\includegraphics[width=1in,height=1.25in,clip,keepaspectratio]
{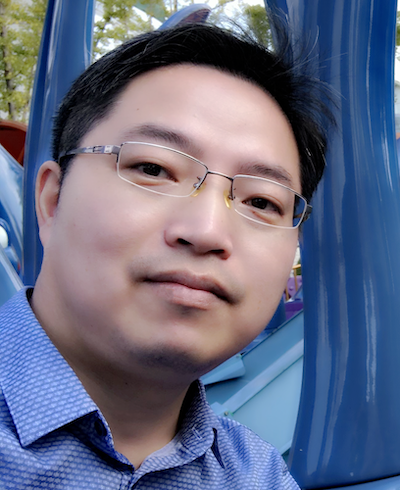}}]{Zhiyong Su}
is currently an associate professor at the School of Automation, Nanjing University of Science and Technology, China. He received the B.S. and M.S. degrees from the School of Computer Science and Technology, Nanjing University of Science and Technology in 2004 and 2006, respectively, and received the Ph.D. from the Institute of Computing Technology, Chinese Academy of Sciences in 2009. His current interests include computer graphics, computer vision, augmented reality, and machine learning.
\end{IEEEbiography}

\end{document}